\documentclass[10pt,twocolumn,letterpaper]{article}

\usepackage[pagenumbers]{iccv} 

\usepackage{multirow}

\usepackage{listings}
\usepackage{epsfig}
\usepackage{verbatim}
\usepackage{booktabs}
\usepackage{tabularx}
\usepackage{amsfonts}
\usepackage{multirow}
\usepackage{bbding}
\usepackage{algorithm,algorithmicx,algpseudocode}
\usepackage{array}
\usepackage{colortbl}
\usepackage{wasysym}
\usepackage{amsmath}
\usepackage{amssymb}
\usepackage{pifont}
\definecolor{mygray}{gray}{.92}
\definecolor{lightgray}{gray}{.96}
\definecolor{myy}{RGB}{126,95,0}
\definecolor{ggray}{RGB}{127,127,127}
\definecolor{mygreen}{RGB}{0,0,0}
\definecolor{myred}{RGB}{240,16,89}
\definecolor{myblue}{RGB}{0,114,188}
\definecolor{darkgreen}{rgb}{0.0, 0.5, 0.0}
\definecolor{demphcolor}{RGB}{100,100,100}

\usepackage{caption}
\makeatletter
\newcommand{\thickhline}{%
    \noalign {\ifnum 0=`}\fi \hrule height 0.8pt
    \futurelet \reserved@a \@xhline
}

\usepackage{etoolbox}
\makeatletter
\AfterEndEnvironment{algorithm}{\let\@algcomment\relax}
\AtEndEnvironment{algorithm}{\kern2pt\hrule\relax\vskip3pt\@algcomment}
\let\@algcomment\relax
\newcommand\algcomment[1]{\def\@algcomment{\footnotesize#1}}
\renewcommand\fs@ruled{\def\@fs@cfont{\bfseries}\let\@fs@capt\floatc@ruled
  \def\@fs@pre{\hrule height.8pt depth0pt \kern2pt}%
  \def\@fs@post{}%
  \def\@fs@mid{\kern2pt\hrule\kern2pt}%
  \let\@fs@iftopcapt\iftrue}
\makeatother
\definecolor{mygray}{gray}{.92}
\definecolor{mygreen}{RGB}{93,173,85}

\makeatother

\newcolumntype{d}[1]{>{\raggedright\arraybackslash}p{#1pt}}
\newcolumntype{e}[1]{>{\raggedleft\arraybackslash}p{#1pt}}

\definecolor{baselinecolor}{gray}{.9}

\newlength\savewidth

\renewcommand{\paragraph}[1]{\vspace{1.25mm}\noindent\textbf{#1}}

\newcolumntype{x}[1]{>{\centering\arraybackslash}p{#1pt}}
\newcolumntype{y}[1]{>{\raggedright\arraybackslash}p{#1pt}}
\newcolumntype{z}[1]{>{\raggedleft\arraybackslash}p{#1pt}}

\newcommand{\app}{\raise.17ex\hbox{$\scriptstyle\sim$}}

\definecolor{deemph}{gray}{0.6}

\definecolor{baselinecolor}{gray}{.9}

\definecolor{color_green}{HTML}{92D050}
\definecolor{iccvblue}{rgb}{0.21,0.49,0.74}
\usepackage[pagebackref,breaklinks,colorlinks,allcolors=iccvblue]{hyperref}

\def\paperID{1517} 
\def\confName{ICCV}
\def\confYear{2025}

\title{MedSegFactory: Text-Guided Generation of Medical Image-Mask Pairs}

\author{%
Jiawei Mao$^{1}$
\quad
Yuhan Wang$^{1}$ \quad
Yucheng Tang$^{2}$ \quad 
Daguang Xu$^2$ \\
Kang Wang$^3$ \quad
Yang Yang$^3$ \quad
Zongwei Zhou$^4$ \quad
Yuyin Zhou$^1$ \vspace{1.5em} \\
$^1$UC Santa Cruz
\quad $^2$NVIDIA \quad $^3$UC San Francisco \quad $^4$Johns Hopkins University
}

\begin{document}

\twocolumn[{%
\renewcommand\twocolumn[1][]{#1}%
\maketitle
\vspace{-3em}
\begin{center}
    Project: ~\href{https://jwmao1.github.io/MedSegFactory_web/}{MedSegFactory}
\end{center}

\begin{center}
\setlength{\abovecaptionskip}{3mm} 
\setlength{\belowcaptionskip}{-4mm}
\centering
\vspace{-0.2cm}
\setlength{\belowcaptionskip}{4mm}
\includegraphics[width=.98\linewidth]{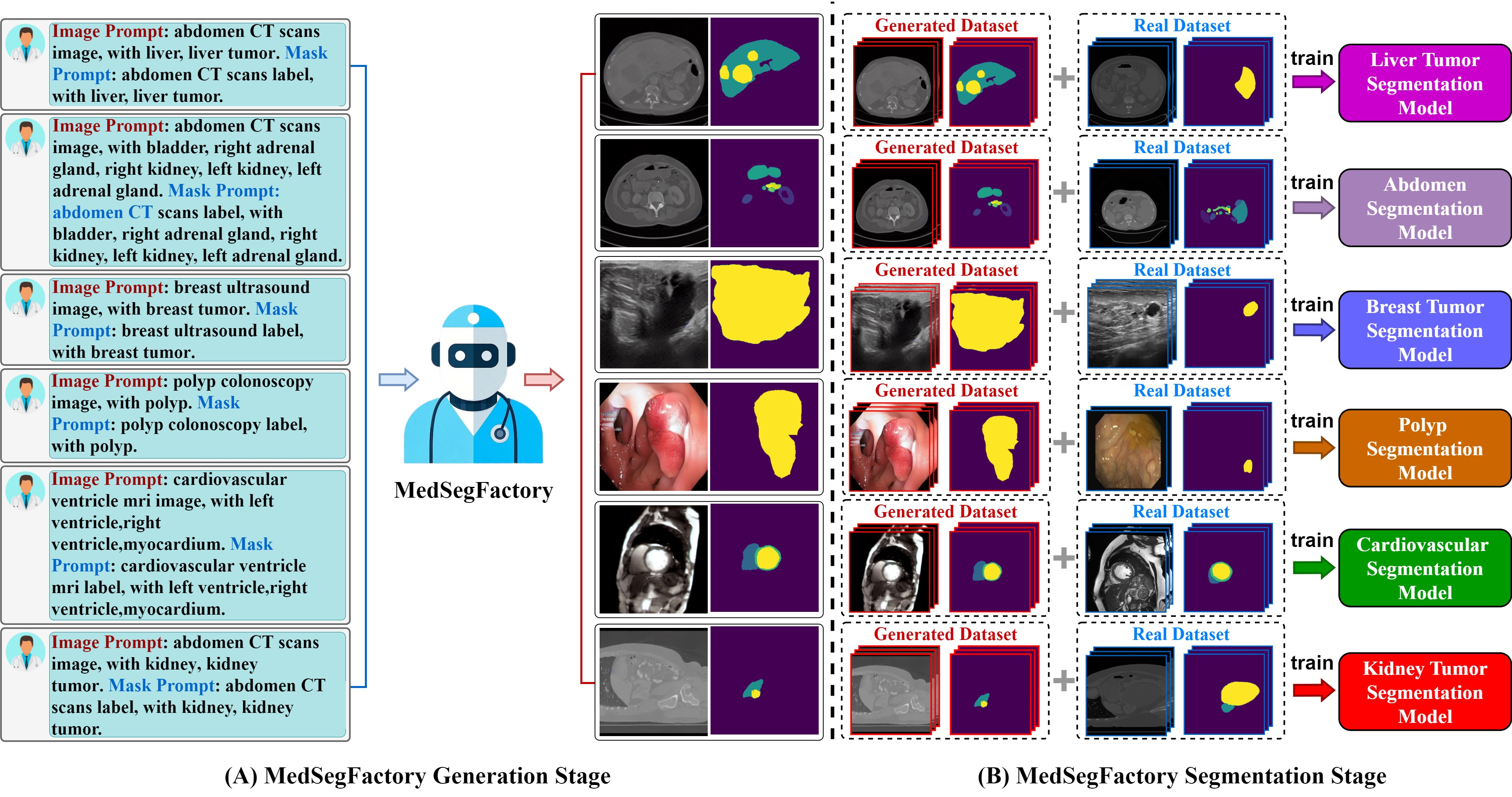}
\vspace{-.8em}
\captionof{figure}{(A) MedSegFactory is a \emph{versatile} medical image synthesis tool that generates paired medical images and segmentation masks \emph{solely from user-defined prompts} specifying the target label and imaging modality.
(B) The synthesized data generated by MedSegFactory can be then used to augment datasets, enhancing segmentation performance across various imaging modalities.} 
\label{fig_teaser}
\end{center}

}]

\begin{abstract}
This paper presents \textbf{MedSegFactory}, a versatile medical synthesis framework that generates high-quality paired medical images and segmentation masks across modalities and tasks. It aims to serve as an unlimited data repository, supplying image-mask pairs to enhance existing segmentation tools. The core of MedSegFactory is a dual-stream diffusion model, where one stream synthesizes medical images and the other generates corresponding segmentation masks. To ensure precise alignment between image-mask pairs, we introduce Joint Cross-Attention (JCA), enabling a collaborative denoising paradigm by dynamic cross-conditioning between streams. This bidirectional interaction allows both representations to guide each other's generation, enhancing consistency between generated pairs.
MedSegFactory unlocks on-demand generation of paired medical images and segmentation masks through user-defined prompts that specify the target labels, imaging modalities, anatomical regions, and pathological conditions, facilitating scalable and high-quality data generation.
This new paradigm of medical image synthesis enables seamless integration into diverse medical imaging workflows, enhancing both efficiency and accuracy. Extensive experiments show that MedSegFactory generates data of superior quality and usability, achieving competitive or state-of-the-art performance in 2D and 3D segmentation tasks while addressing data scarcity and regulatory constraints. 
\end{abstract}    
\section{Introduction}\label{sec:intro}

Computer-aided detection systems~\cite{ronneberger2015u,isensee2021nnu,kirillov2023segment} have significantly enhanced medical image analysis, but their widespread adoption remains limited due to the high costs associated with accessing and annotating medical imaging data~\cite{kang2023label,liu2023clip,li2024abdomenatlas,zhou2019semi,zhou2019high,zhou2019prior}. Annotated datasets are costly to create and are restricted by strict privacy and regulatory policies, which hinder data sharing between institutions. Synthetic data present a promising alternative by creating large-scale annotated datasets without compromising patient confidentiality, enabling easier collaboration among researchers.

\begin{figure}[t]
    \setlength{\abovecaptionskip}{3mm} 
    \setlength{\belowcaptionskip}{-4mm}
    \centering
    \includegraphics[width=1\linewidth]{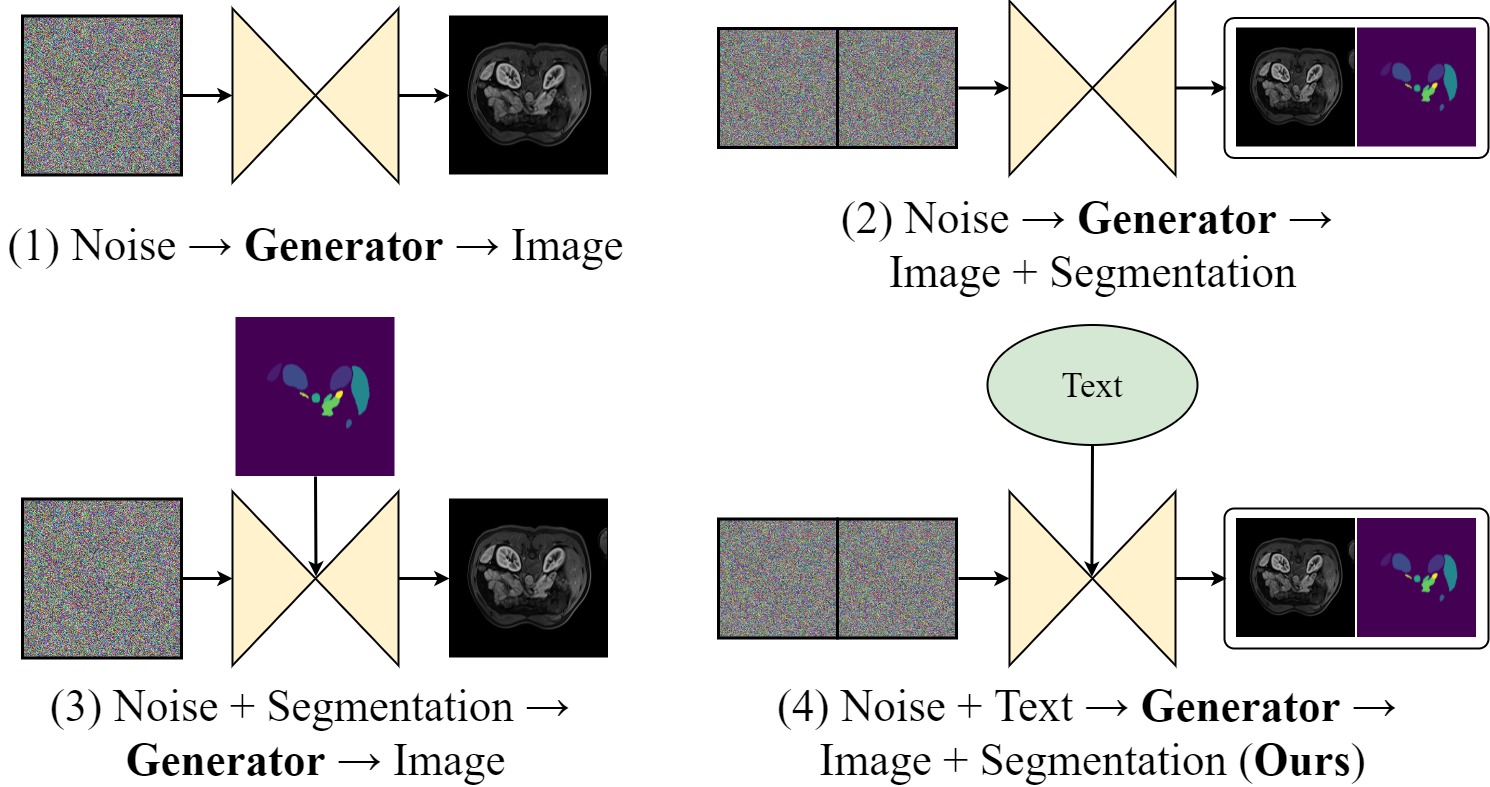}
     \caption{\textbf{Different medical image generation frameworks:} (1) unconditional image generation, (2) unconditional paired image-mask generation, and (3) conditional image generation guided by segmentation masks. (1) and (2) lack control, while (3) depends on costly and hard-to-obtain mask annotations, limiting scalability. 
     }
     \label{fig_task}
\end{figure}

Existing methods for medical data synthesis~\cite{chen2024analyzing,goodfellow2020generative,brock2018large,ho2020denoising,song2020score,Karras2019stylegan2} generally fall into three categories, as summarized in Fig.~\ref{fig_task}: (1) unconditional image generation~\cite{ho2020denoising,song2020score,rombach2022high,nichol2021improved,zhang2024diffboost,dorjsembe2024conditional,friedrich2024wdm}, (2) unconditional paired image-mask generation~\cite{sams2022gan,zhu2017unpaired,machavcek2023mask}, and (3) conditional image generation guided by segmentation masks~\cite{zhang2023adding,qadir2022simple,thambawita2022singan,dorjsembe2024conditional,dorjsembe2024polyp,du2023arsdm}. Each of these categories, however, has critical limitations. Unconditional methods offer limited control, complicating interpretability, debugging and resulting in images with reduced variability. Conditional methods provide greater control and interpretability, but they depend heavily on detailed mask annotations, which are expensive and difficult to obtain, limiting scalability.

To address these fundamental limitations, we introduce a new, fourth category of data synthesis: conditional paired image-mask generation based on text prompts. Within this category, we introduce MedSegFactory, a versatile medical synthesis framework designed to generate medical images and their corresponding segmentation masks only from concise textual descriptions. Unlike previous conditional methods~\cite{du2023arsdm,qadir2022simple,guo2024maisi,wu2024mrgen}, users no longer need to provide detailed segmentation masks; instead, they simply provide a brief text prompt describing high-level features such as target labels, imaging modalities, anatomical regions, and pathological conditions. This approach significantly simplifies the creation of customized synthetic datasets.

To equip MedSegFactory with broad generative capability across diverse medical imaging tasks, the model is trained on multiple image modalities and tasks. At its core, MedSegFactory employs a dual-stream diffusion model, where one stream synthesizes medical images while the other generates corresponding segmentation masks. To ensure precise alignment between image-mask pairs, we introduce Joint Cross-Attention, leveraging the cross-attention interactions between the image and mask streams, dynamically conditioning each stream's output on the evolving denoising representations of the other. This mutual, interactive refinement ensures robust semantic alignment, significantly enhancing the quality and usability of the generated data.

The key contribution of this paper is the MedSegFactory, a valuable tool for advancing medical imaging. MedSegFactory can accurately leverage user instructions (Fig.~\ref{fig:diversity}; alignment between text and generated image-mask pairs), synthesizing \underline{\emph{high-quality}} (Fig.~\ref{fig_total_imgs}, Tabs.~\ref{tab:3D_vis}--\ref{tab:2D_vis}; comparable or state-of-the-art medical image generation), \underline{\emph{diverse}} (Fig.~\ref{fig:diversity}; varied outputs from the same text prompt), and \underline{\emph{highly consistent}} medical image-segmentation pairs (Fig.~\ref{fig:align}).
This capability significantly enhances the overall segmentation performance across heterogeneous datasets (Tabs.~\ref{tab:seg_other}--\ref{tab:seg_ct}; consistently achieving comparable or state-of-the-art performance in downstream medical segmentation tasks).

Overall, our contributions can be summarized as follows:

\begin{enumerate}
    \item We introduce MedSegFactory, a dual-stream generative model designed to synthesize highly consistent image-mask pairs via text prompts. 
    
    \item We propose a novel Joint Cross-Attention Mechanism, which enables mutual refinement between the image and mask streams, ensuring improved semantic consistency and precise alignment in the generated pairs. 
    
    \item The synthetic data generated by MedSegFactory demonstrate comparable or superior quality and usability to previous state-of-the-art methods, as evidenced by medical image generation and downstream 2D \& 3D segmentation tasks (Sec.~\ref{sec:exp_seg}).
\end{enumerate}

\section{Related Work}\label{sec:related}

\paragraph{Diffusion Model}
Diffusion models learn data distributions by progressively denoising noisy inputs through a series of iterative steps. Specifically, DDPM~\cite{ho2020denoising} formulates diffusion models as a more stable generative framework by designing a loss function based on a Markov chain, effectively overcoming the limitations of Generative Adversarial Networks (GANs)~\cite{goodfellow2020generative}. Designs such as DDIM~\cite{song2020denoising} and LCM~\cite{luo2023latent} improve the sampling efficiency issue of the DDPM framework. Stable Diffusion Model (SDM)~\cite{rombach2022high} explores the application of diffusion models on the latent space. Building upon the high-fidelity image generation framework of SDM, ControlNet~\cite{zhang2023adding} introduces various conditional controls, such as Canny edge detection, HED boundary detection, and semantic segmentation, to guide image generation in diffusion models. Due to their high-fidelity generation capabilities and precise conditional control, diffusion models have been successfully applied to various downstream tasks, including medical image generation~\cite{polamreddy2024leapfrog,du2023arsdm,dorjsembe2024conditional,friedrich2024wdm,chen2024towards,zhang2023self,li2024text}, video synthesis~\cite{esser2023structure,lee2024grid,khachatryan2023text2video,jiang2025lung}, and low-level vision applications~\cite{yi2023diff,wang2024image,ai2024multimodal}. However, generating paired medical images and segmentation labels remains a challenge for diffusion models due to privacy concerns and the high cost of annotation.

\paragraph{Medical Image Generation}
Synthetic medical images can augment limited medical databases and help mitigate data scarcity resulting from privacy and ethical constraints.  
However, due to the absence of corresponding medical segmentation labels, methods~\cite{nie2018medical,dalmaz2022resvit,skandarani2023gans,cho2024medisyn,hagan2022} that generate only medical images are inadequate for supporting downstream computer-aided detection (CAD). Medical image generation schemes~\cite{du2023arsdm,qadir2022simple,guo2024maisi,wu2024mrgen} based on medical segmentation labels can generate clinically relevant images suitable for CAD. However, the limited diversity of available medical segmentation labels hinders advancements in CAD, as the high cost of manual annotation remains a significant challenge. Although \citet{bhat2025simgen,machavcek2023mask} address the aforementioned issues, their framework design remains limited in applicability to specific medical image generation tasks and corresponding CAD applications.

\paragraph{Medical Segmentation} 
As a fundamental task in CAD, medical segmentation automatically delineates lesion areas, providing precise structural information for surgical planning and detection. Conventional segmentation frameworks, such as FCN~\cite{long2015fully} and U-Net~\cite{ronneberger2015u}, demonstrated notable effectiveness in CAD by leveraging fully convolutional networks and skip connections, respectively. U-Net system~\cite{isensee2021nnu,schlemper2019attention,huang2020unet,ruan2024vm,yan2022after} adaptively incorporates different network structures and sample sizes according to specific requirements to better respond to diverse scenarios and challenges. However, the limited receptive field of convolutional networks restricts their ability to effectively capture long-range dependencies and inter-region relationships within complex structures, such as lesions or organs. Thus, numerous studies~\cite{chen2021transunet,perera2024segformer3d,azad2023enhancing,azad2022transdeeplab,huang2021missformer,cao2022swin,zhang2021transfuse} have attempted to integrate Transformers into medical segmentation frameworks to globally model lesion or organ representations. But these efforts focus only on particular medical segmentation tasks. Additionally, unified frameworks have been designed~\cite{ma2024segment,wang2024fsam,ren2024medical} to address discrepancies across different medical datasets and tasks. However, the scarcity of medical data has increasingly become a significant barrier to the continued advancement of unified segmentation.
\begin{figure*}[!t]
    \setlength{\abovecaptionskip}{-4mm} 
    \setlength{\belowcaptionskip}{-4mm}
    \centering
    \includegraphics[width=1\linewidth]{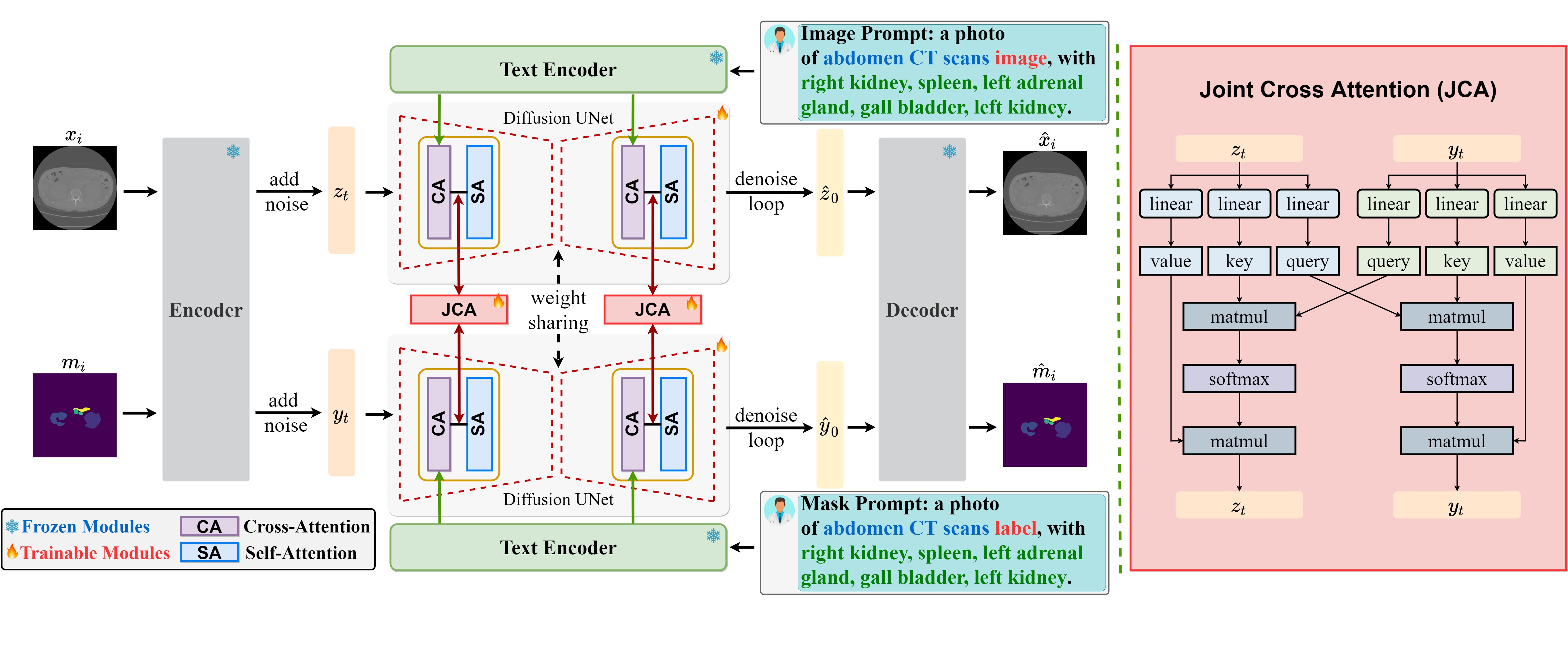}
     \caption{\textbf{Overview of the MedSegFactory Training Pipeline.} \emph{Left:} A dual-stream generation framework built upon the Stable Diffusion model, where each paired image and segmentation mask \(\{x_{i}, m_{i}\}\) are first encoded into latent representations and noised. These noised representations are then denoised by the Diffusion U-Net, producing the image and segmentation mask latent \(\{\hat{z}_{0}, \hat{y}_{0}\}\), which are ultimately decoded into a paired image and mask $\{\hat x_{i},\hat m_{i}\}$. To improve their semantic consistency, we introduce \textbf{Joint Cross-Attention (JCA)}, which allows the denoised latent representations of both streams to act as dynamic conditions, enhancing the mutual generation of images and masks. \emph{Right:} A detailed illustration of JCA, which applies two cross-attention functions to the intermediate denoising features \(\{z_{t}, y_{t}\}\), enabling the cross-conditioning of the medical image and segmentation label for improved consistency in the final outputs.}
     \label{fig_framework}
   \end{figure*}

\section{Method}\label{sec:method}
In this paper, we introduce MedSegFactory, a novel medical image synthesis framework that generates paired medical images and segmentation masks based on user-defined prompts.
We first define this synthesis task in Sec.~\ref{sec:task}, then outline our training pipeline (Fig.~\ref{fig_framework}) for developing a versatile model capable of synthesizing medical images across multiple modalities and tasks.
Sec.~\ref{sec:framework} details MedSegFactory’s dual-stream generation framework, where one stream synthesizes images while the other produces corresponding masks.
To ensure semantic alignment, we propose a Joint Cross-Attention (JCA) mechanism (Sec.~\ref{sec:jca}) that enables bidirectional conditioning between streams, allowing dynamic refinement and ensuring consistency between generated image-mask pairs.

\subsection{Task Definition}\label{sec:task} 
We define a new medical image synthesis task that turns a user-defined prompt—specifying the target label and imaging modality—into the unified generation of paired medical images and segmentation labels.


\paragraph{Generation.} Given a real dataset $S_{r}=\{\{x_{i},p_{i},m_{i}\}\}_{i=1}^{n}$, where $n$ is the data size in $S_{r}$, $x_{i}\in \mathbb{R}^{3\times h\times w}$ denotes a medical image (with height $h$ and width $w$), $p_{i}$ denotes the corresponding text prompt, and $m_{i} \in \mathbb{R}^{h\times w}$ is the paired segmentation mask, our generative function $f_{g}$ can be defined as:
\begin{equation}
   \begin{split}
         &\{\hat x_{i},\hat m_{i}\} \leftarrow f_{g}(p_{i}).
  \end{split}
   \label{eq:task}
 \end{equation}
where $\{\hat x_{i},\hat m_{i}\}$ are the synthesized medical image and its corresponding segmentation mask.

\paragraph{Segmentation.} For medical segmentation tasks, we use both real $D_{r}=\{\{x_{i},m_{i}\}\}_{i=1}^{n}$ and generated datasets $D_{g}=\{\{\hat x_{i},\hat m_{i}\}\}_{i=1}^{n}$ to jointly train the segmentation function $f_{s}$.
\begin{equation}
   \begin{split}
         &\mathring m_{i}, \overline m_{i} \leftarrow f_{s}(x_{i}), f_{s}(\hat x_{i}).
  \end{split}
   \label{eq:task}
 \end{equation}
where $\mathring m_{i}$ and $\overline m_{i}$ are the predicted segmentation labels for $x_{i}$ and $\hat x_{i}$ respectively. The objective of $f_{s}$ is to maximize the likelihood that the predicted labels $\{\mathring m_{i}, \overline m_{i}\}$  align with their corresponding ground truth labels $\{m_{i}, \hat m_{i}\}$.

\subsection{Dual-Stream Generation}\label{sec:framework} 
As illustrated in Fig.~\ref{fig_framework}, MedSegFactory is built upon a dual-stream generation framework, from the Stable Diffusion model~\cite{rombach2022high}. 
The framework consists of two parallel streams: one for generating medical images and the other for generating the corresponding segmentation masks.  

During training, each paired medical image and segmentation mask \((x_{i}, m_{i})\) is first encoded into latent representations \(\{z_{0}, y_{0}\}\) using a pretrained variational autoencoder (VAE) encoder. 
At the \(t\)-th denoising step, the intermediate latent representations \(\{z_{t}, y_{t}\}\) are obtained by applying Gaussian noise \(\{\epsilon_x, \epsilon_m\}\) to the initial encoded vectors \(\{z_{0}, y_{0}\}\), the forward diffusion process:  
\begin{equation}
   \begin{split}
    &q(z_t, y_t \mid z_0, y_0) = \mathcal{N}(z_t; \sqrt{\alpha_t} z_0, 1 - \alpha_t)\\ 
    &\cdot \mathcal{N}(y_t; \sqrt{\alpha_t} y_0, 1 - \alpha_t), 
  \end{split}
   \label{eq:forward_diff}
 \end{equation}  
where \(\alpha_t\) represents the noise scaling factor at step \(t\).  
 
$z_{t}$ and $y_{t}$ are then denoised through a Diffusion UNet. Specifically, as show in Fig.~\ref{fig_framework}, the denoising process is performed in the Cross-Attention function $CA(Q,K,V)$ (where $Q,K,V$ are the queries, keys, and values of cross-attention respectively) and self-attention $SA(Q,K,V)$ relying on the corresponding textual prompt embeddings $e_{x}$ and $e_{m}$ obtained by fixed text encoder, respectively: 
 \begin{equation}
   \begin{split}
    &z_{t},y_{t}=CA(z_{t},e_{x},e_{x}),CA(y_{t},e_{m},e_{m}), \\
    &z_{t},y_{t}=SA(z_{t},z_{t},z_{t}),SA(y_{t},y_{t},y_{t}).
  \end{split}
   \label{eq:forward_diff}
 \end{equation}

Finally, the denoised latent representations are decoded by the VAE decoder into a medical image $\hat x_{i}$ and its corresponding segmentation mask $\hat m_{i}$.

Unlike traditional conditional generation models~\cite{zhang2023adding,guo2024maisi,du2023arsdm}, which rely on predefined segmentation masks as static conditional inputs, our approach assumes that segmentation masks are not readily available. MedSegFactory instead introduces a dynamic denoising process enabled by our newly proposed Joint Cross-Attention (JCA) mechanism. This mechanism, detailed below, allows both medical images and segmentation masks to be progressively refined based on each other’s denoised representations, ensuring mutual enhancement throughout the generation process.

\subsection{Joint Cross-Attention}\label{sec:jca}
Despite the success of diffusion models~\cite{ho2020denoising,rombach2022high,song2020score} in generative modeling, their application to jointly modeling the correspondence between medical images and segmentation labels remains underexplored. Unlike existing frameworks that generate medical images from predefined segmentation labels, MedSegFactory models the joint distribution of medical images and their corresponding segmentation masks, ensuring mutual refinement during the generation process.

\begin{figure*}[!t]
    \setlength{\abovecaptionskip}{3mm} 
    \setlength{\belowcaptionskip}{2mm}
    \centering
    \includegraphics[width=1\linewidth]{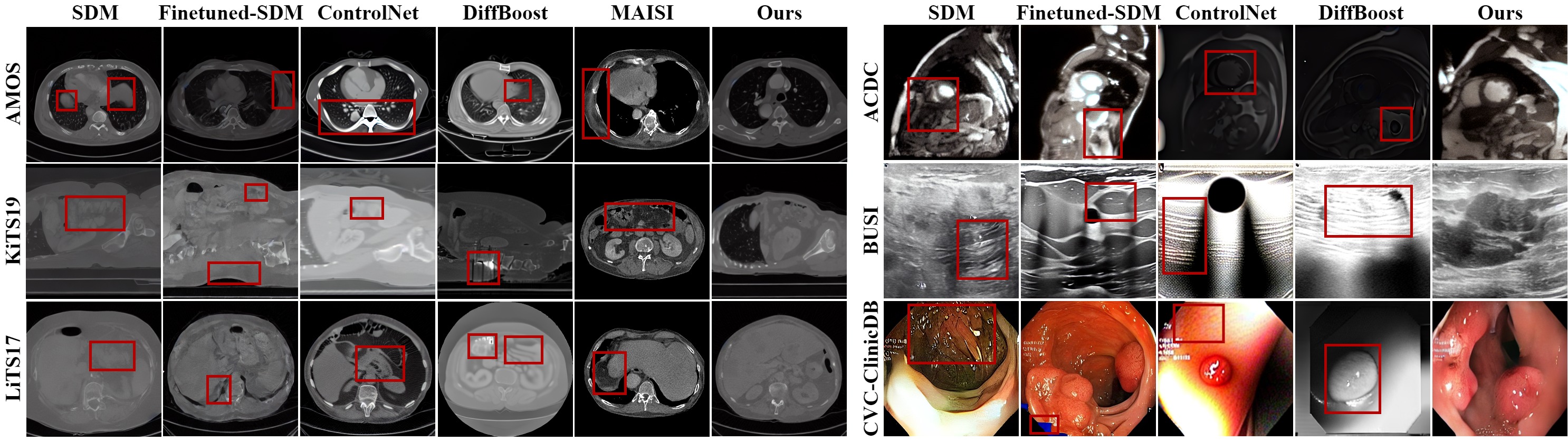}
     \caption{\textbf{Qualitative comparison between MedSegFactory and baselines for medical image generation.} Baselines exhibit issues: irregular structures (SDM, ControlNet, DiffBoost, Finetuned-SDM), artifacts (SDM, DiffBoost, Finetuned-SDM), blurred boundaries (Finetuned-SDM), unnatural saturation (ControlNet, DiffBoost), and structural missing (MAISI). MedSegFactory produces anatomically accurate, high-quality images. Zoom in for details.}
     \label{fig_total_imgs}
          \vspace{-0.6cm}
   \end{figure*}

This can be achieved through our Joint Cross-Attention (JCA) 
mechanism, which enhances the UNet backbone of the diffusion model, making it more adaptable for paired data generation. 
Specifically, to capture the correspondence between medical images and segmentation labels, JCA applies two cross-attention functions to intermediate denoising features, enabling mutual refinement:
\begin{equation}
   \begin{split}
    &z_{t},y_{t}=CA(y_{t},z_{t},z_{t}), CA(z_{t},y_{t},y_{t}). 
  \end{split}
   \label{eq:forward_diff}
 \end{equation}
Unlike conventional single-stream conditional generation models, which rely on static segmentation labels, MedSegFactory employs JCA to dynamically adjust both the image and label representations throughout the denoising process. 
This collaborative denoising mechanism facilitates bidirectional interaction between latent representations, enabling mutual refinement at each step, and thus ensures stronger semantic consistency between generated images and masks.

Formally, the overall objective of MedSegFactory can be summarized as follows:
\begin{equation}
 \begin{split}
    \mathbb{E}_{t, \epsilon_x, \epsilon_m, e_{x}, e_{m}} 
    &[\left\| \epsilon_x - \epsilon_{\theta}(z_t, y_t, t, e_x) \right\|_1 \\
    &+ \left\| \epsilon_m - \epsilon_{\theta}(y_t, z_t, t, e_m) \right\|_1].
 \end{split}
 \label{eq:obj}
\end{equation}
where $\epsilon_{\theta}$ is the neural backbone of MedSegFactory. 

\section{Experiment}\label{sec:experiment} 
To demonstrate the effectiveness of MedSegFactory, we show the quality of images generated by MedSegFactory in Sec.~\ref{sec:exp_gen}. We investigate the generated results of MedSegFactory in medical image segmentation tasks (Sec.~\ref{sec:exp_seg}).

\subsection{Dataset}

\begin{table}[ht!]
\setlength{\abovecaptionskip}{1mm} 
\setlength{\belowcaptionskip}{-2mm}
\small
\resizebox{1.00\columnwidth}{!}{
\begin{tabular}{l|l|l|rr}
            \textbf{Dataset}  & \textbf{Modalities}  & \textbf{Classes} & \textbf{\#Train} & \textbf{\#Test} \\
            \toprule
            \multirow{1}{*}{AMOS}~\cite{ji2022amos} & CT Scan & 16 & 35524 & 18537 \\
            \hline
            \multirow{1}{*}{ACDC}~\cite{bernard2018deep} & MRI & 4 & 1902 & 3826 \\
            \hline
            \multirow{1}{*}{BUSI}~\cite{al2020dataset} & Ultrasound Scan & 2 & 547 & 233 \\
            \hline
            \multirow{1}{*}{CVC-ClinicDB}~\cite{bernal2015wm} \& {Kvasir-SEG}~\cite{jha2020kvasir} & Colonoscopy & 2 & 1450 & 62 \\
            \hline
            \multirow{1}{*}{KiTS19}~\cite{heller2023kits19} & CT Scan & 3 & 97052 & 10752 \\
            \hline
            LiTS17~\cite{bilic2023liver} & CT Scan & 3 & 55815 & 2823 \\
            \bottomrule
        \end{tabular}
}
\caption{Dataset details for training and evaluation.}
\label{tab_dataset}
\end{table}

To validate MedSegFactory generative results for paired medical images and segmentation labels, we collected seven publicly available medical segmentation datasets, includes multiple modalities from 2D and 3D databases: AMOS, ACDC, BUSI, CVC-ClinicDB, KiTS19, and LiTS17. 
We trained MedSegFactory and baselines using the training partitions of each dataset, including 2D slices or images from either the axial view of the whole CT scan. The test partitions were used for evaluating both the generation and downstream tasks (see Details in Tab.~\ref{tab_dataset}).

\begin{figure*}[!t]
    \setlength{\abovecaptionskip}{1mm} 
    \setlength{\belowcaptionskip}{-4mm}
    \centering
    \includegraphics[width=1\linewidth]{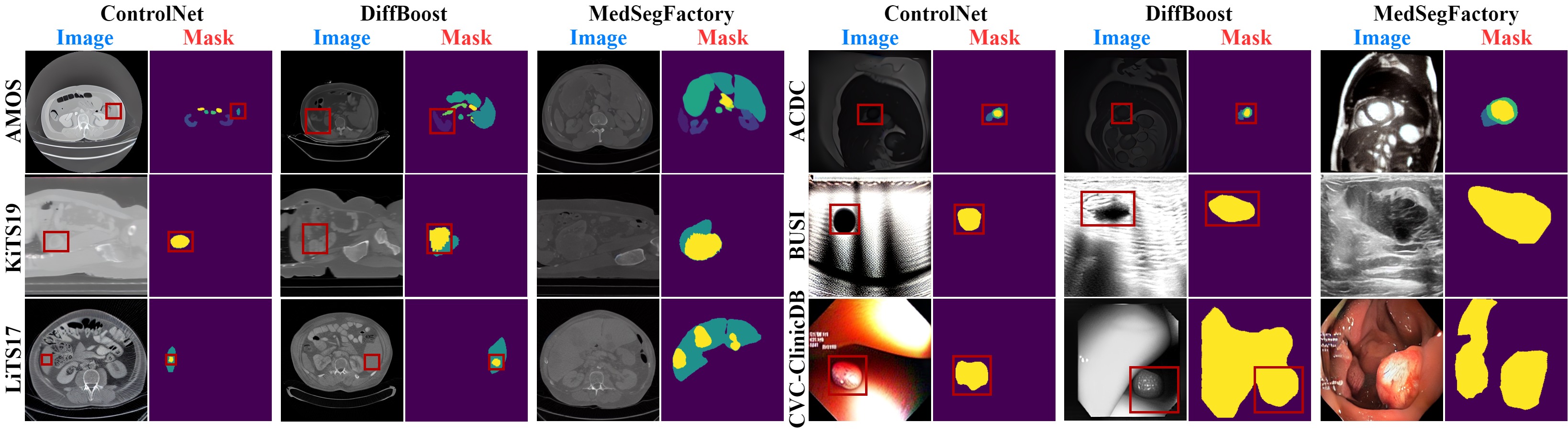}
     \caption{\textbf{Alignment comparison} reveals misalignment or detail loss in baseline-generated medical images versus segmentation labels (highlighted in \textcolor{red}{red box}). MedSegFactory demonstrates superior alignment and finer details. 
     }
     \label{fig:align}
   \end{figure*}

\subsection{Implementation Details and Metrics}

\paragraph{Implementation Details.} We implemented MedSegFactory following the Stable Diffusion Model~\cite{rombach2022high} (SDM) architecture configuration. The model was trained for 300 epochs using the AdamW optimizer with a learning rate of 1$e$-5 and a batch size of 20. Training was conducted on 4 NVIDIA A6000 GPUs using PyTorch. To accommodate the unbalanced aspect ratios of medical slices, we follow SDXL's Multi-Aspect Training~\cite{podell2023sdxl} while preserving a dimension close to $256^{2}$ pixels. We employed classifier-free guidance with a scale of 7.5 for inference, generating $256^{2}$ pixels medical images and corresponding segmentation labels through a 50-step denoising process. We evaluated at 256 resolution upon their original settings for baselines.

\paragraph{Evaluation Metrics.} For medical image generation, we use Inception Score (IS) and Fréchet Inception Distance (FID) as evaluation metrics. For segmentation, Dice coefficients (DSC) and Intersection over Union (IoU) are used to compare the performance of MedSegFactory and baselines.

\subsection{Medical Image Generation}\label{sec:exp_gen}
To assess the generation quality of MedSegFactory, we compare it against previous state-of-the-art methods ControlNet~\cite{zhang2023adding}, MAISI~\cite{guo2024maisi}, Stable Diffusion Model~\cite{rombach2022high}, fine-tuned SDM~\cite{rombach2022high}, and DiffBoost~\cite{zhang2024diffboost}. Since MAISI is designed only for 3D image generation, we use its pre-trained weights from large public datasets, while the other baselines are trained on the same training set as MedSegFactory. For each method, we generate a set of images that matches the size of the corresponding test set. For fair comparison, 3D volumes from MAISI are sliced into 2D images.

\begin{table}[ht!]
\setlength{\abovecaptionskip}{1mm} 
\setlength{\belowcaptionskip}{-2mm}
\small
\resizebox{1.00\columnwidth}{!}{
\begin{tabular}{c|cc|cc|cc}
\hline
\multirow{2}{*}{Method} & \multicolumn{2}{c|}{AMOS} & \multicolumn{2}{c|}{KiTS19} & \multicolumn{2}{c}{LiTS17}\\ \cline{2-7}
& FID $\downarrow$ & IS $\uparrow$ & FID $\downarrow$ & IS $\uparrow$ & FID $\downarrow$ & IS $\uparrow$\\ \hline\hline
SDM~\cite{rombach2022high}  & 98.11 & 3.58 & 91.16 & 2.65 & 105.58 & 2.48\\ \hline
Finetuned-SDM~\cite{rombach2022high} & 91.50 & 3.19 & 106.70 & 3.07 & 104.11 & 2.46 \\ \hline
MAISI~\cite{guo2024maisi}  & 158.79 & 3.09  & 161.59 & 2.29 & 175.73 & 2.25\\ \hline
ControlNet~\cite{zhang2023adding} & 105.38 & 3.40 & 117.55 & 2.14 & 132.36 & 2.45 \\ \hline
DiffBoost~\cite{zhang2024diffboost} & \textbf{73.61} & 3.21 & \textbf{84.66} & \textbf{3.18} & 99.32 & \textbf{2.81}\\ \hline
MedSegFactory & 97.94 & \textbf{3.62} & 95.87 & 2.46 & \textbf{91.41} & 2.52\\ \hline
\end{tabular}
}
\caption{Performance comparison of Inception Score and Fréchet Inception Distance for the generated images from MedSegFactory and the baselines in abdominal CT generation tasks.}
\label{tab:3D_vis}
\end{table}

\begin{table}[ht!]
\setlength{\abovecaptionskip}{1mm} 
\setlength{\belowcaptionskip}{-5mm}
\small
\resizebox{0.99\columnwidth}{!}{
\begin{tabular}{c|cc|cc|cc}
\hline
\multirow{2}{*}{Method} & \multicolumn{2}{c|}{ACDC} & \multicolumn{2}{c|}{BUSI} & \multicolumn{2}{c}{CVC-ClinicDB}\\ \cline{2-7}
& FID $\downarrow$ & IS $\uparrow$ & FID $\downarrow$ & IS $\uparrow$ & FID $\downarrow$ & IS $\uparrow$\\ \hline\hline
SDM~\cite{rombach2022high}  & 160.98 & 3.47 & 168.71 & 2.08 & 235.19 & 2.24 \\ \hline
Finetuned-SDM~\cite{rombach2022high} & 165.19 & 3.35 & 159.74 & \textbf{2.83} & 246.55 & 2.39\\ \hline
ControlNet~\cite{zhang2023adding} & 173.77 & 2.52 & 170.50 & 2.38 & 250.24 & 2.31 \\ \hline
DiffBoost~\cite{zhang2024diffboost} & 198.36 & 2.37 & 184.85 & 2.33 & 286.29 & 2.08\\ \hline
MedSegFactory & \textbf{158.11} & \textbf{3.51} & \textbf{156.97} & 2.44 & \textbf{227.88} & \textbf{2.43} \\ \hline
\end{tabular}
}
\caption{Performance comparison of Inception Score and Fréchet Inception Distance for the generated images from MedSegFactory and the baselines on ACDC, BUSI, and CVC-ClinicDB.}
\label{tab:2D_vis}
\end{table}

\paragraph{Quantitative Comparison.}
We evaluated the differences between the generated images and real 2D medical slices in Tabs.~\ref{tab:3D_vis}\&\ref{tab:2D_vis}, respectively. As shown in Tab.~\ref{tab:3D_vis}, since DiffBoost~\cite{zhang2024diffboost} employs the HED Boundary of medical images as a synthesis condition, it exhibits state-of-the-art (SoTA) IS and FID on some abdomen CT medical image generation tasks. Nonetheless, MedSegFactory still achieved SoTA or comparable generation quality to most baselines on the liver, kidney, and other abdomen CT generation tasks. Meanwhile, we provide comparison results between MedSegFactory and baselines on more generative tasks in Tab.~\ref{tab:2D_vis}. Generated medical images from MedSegFactory exhibit significantly better fidelity compared to baselines. MedSegFactory significantly improves FID and IS, although generated results still showed certain differences from the real data distribution. Note that since MAISI does not involve 2D medical image and ACDC-related generation tasks, it was not included in the comparison in Tab.~\ref{tab:2D_vis}. Collectively, these results demonstrate the potential of MedSegFactory to deliver high-fidelity medical images for a wide variety of medical tasks.

\paragraph{Qualitative Comparison.}
Fig.~\ref{fig_total_imgs} provides a visual comparison of MedSegFactory with baselines across datasets. As shown in the \textcolor{red}{red} rectangular region of Fig.~\ref{fig_total_imgs}, SDM~\cite{rombach2022high}, ControlNet~\cite{zhang2023adding}, and DiffBoost~\cite{zhang2024diffboost} produce irregular medical image structures in some tasks, with visible artifacts appearing in the generated medical images. Moreover, ControlNet shows abnormal saturation in some generation tasks, while DiffBoost fails to preserve the color balance in colonoscopy image generation. Due to the domain gap between natural and medical images, the fine-tuned version of SDM, introducing natural image weights, results in blurred boundary details in CT image generation. Although MAISI~\cite{guo2024maisi} generates high-fidelity medical images, some structural inconsistencies or deficiencies remain. By comparison, MedSegFactory not only generates anatomically precise and plausible structures but also consistently preserves the quality in medical image generation.

\begin{figure*}[!t]
    \setlength{\abovecaptionskip}{0mm} 
    \setlength{\belowcaptionskip}{-4mm}
    \centering
    \includegraphics[width=1\linewidth]{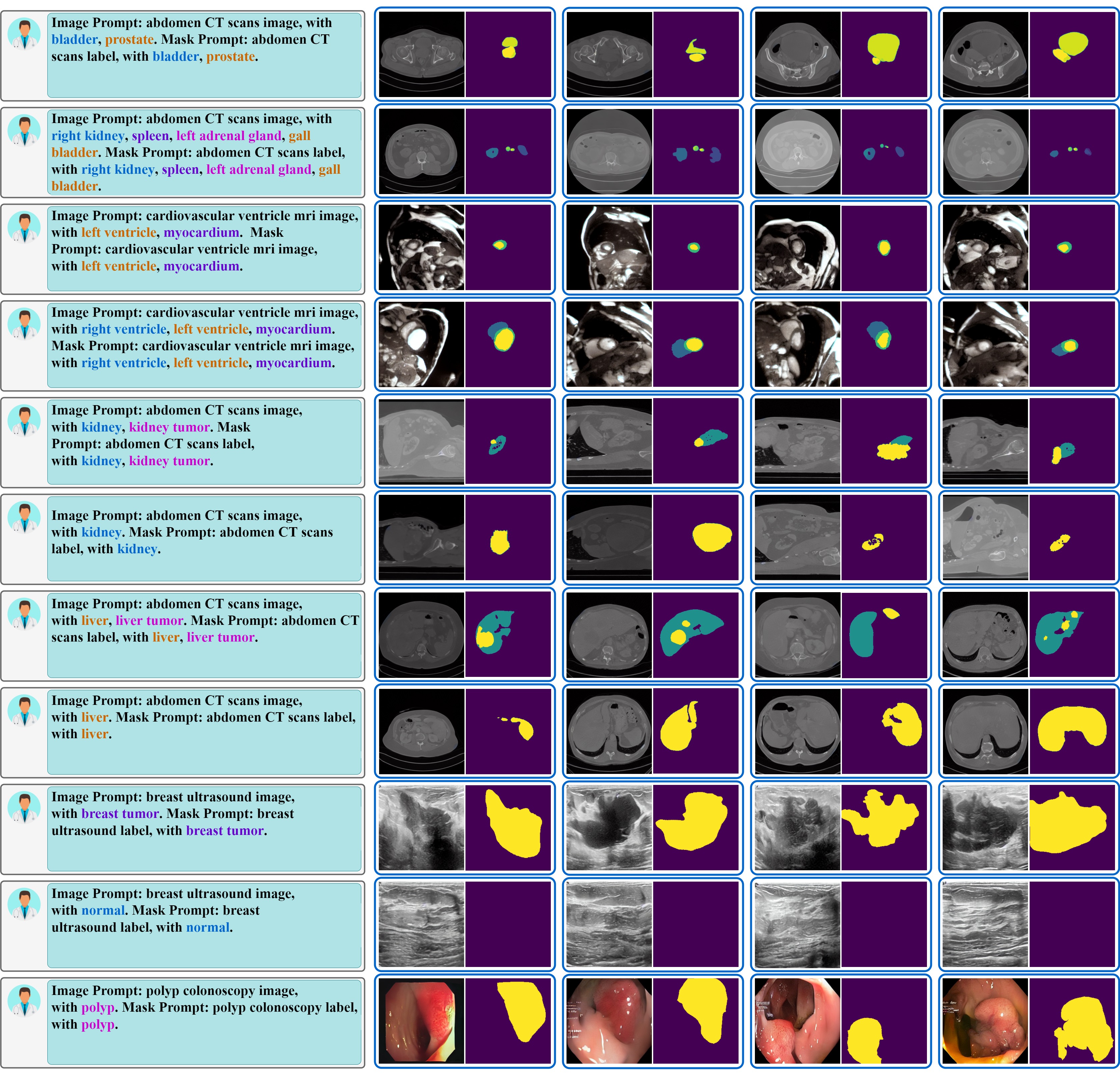}
     \caption{\textbf{MedSegFactory produces high-quality paired medical images and segmentation labels}, precisely following given instructions while demonstrating diversity across multiple outputs under the same conditions.}
     \label{fig:diversity}
   \end{figure*}

\paragraph{Alignment.}
The visualization in Fig.~\ref{fig:align} reveals discrepancies or missing details between the generated medical images and the segmentation labels of the red rectangular regions for both ControlNet and DiffBoost. In contrast, the paired medical images and segmentation labels generated by MedSegFactory exhibit flexible detail alignment and consistency on various tasks.

\paragraph{Diversity.}
Generating a diverse dataset facilitates a wide range of anatomical variations and disease manifestations, which is critical for enhancing the robustness and generalizability of CAD models across various clinical scenarios. Fig.~\ref{fig:diversity} presents multiple pairs of generated results produced by MedSegFactory under the same textual prompt across different generation tasks. As shown in Fig.~\ref{fig:diversity}, the results generated by MedSegFactory accurately align with the provided textual prompts, while the multiple pairs of generated medical images and segmentation labels exhibit semantic consistency. Our paired generation results exhibit sufficient diversity in both medical images and segmentation. 

\subsection{Medical Segmentation}\label{sec:exp_seg}
This section evaluates the contribution of the synthesized medical data for the downstream task of medical segmentation, using the well-established nnUNet~\cite{isensee2021nnu}.

\paragraph{Setting.}
To ensure a fair comparison, the trained MedSegFactory and comparison methods provide synthetic data in the same proportion. The generated medical images and their corresponding segmentation labels are combined to train nnUNet. The testing is conducted exclusively on the held-out test set.

\begin{table}[t]
\setlength{\abovecaptionskip}{1mm} 
\setlength{\belowcaptionskip}{-6mm}
\small
\resizebox{1.00\columnwidth}{!}{
\begin{tabular}{c|cc|cc|cc}
\hline
\multirow{2}{*}{Method} & \multicolumn{2}{c|}{ACDC} & \multicolumn{2}{c|}{BUSI} & \multicolumn{2}{c}{CVC-ClinicDB}\\ \cline{2-7}
& DSC($\%$)  & IoU($\%$)  & DSC($\%$)  & IoU($\%$)  & DSC($\%$) & IoU($\%$) \\ \hline\hline
nnUNet~\cite{isensee2021nnu} & 86.88& 80.42 & 78.77 & 72.73 & 89.45 & 84.72\\ \hline
ControlNet~\cite{zhang2023adding}  & 85.97 & 79.27 & 82.94 & 78.44 & \textbf{91.11} & 86.36\\ \hline
DiffBoost~\cite{zhang2024diffboost} & 87.29 & 80.66 & 78.11 & 72.46 & 89.33 & 84.63\\ \hline
MedSegFactory & \textbf{88.02} & \textbf{81.50} & \textbf{83.38} & \textbf{78.75} & 90.85 & \textbf{86.61}\\ \hline
\end{tabular}
}
\caption{Segmentation performance evaluation of nnUNet on different combined and real datasets.}
\label{tab:seg_other}
\end{table}

\begin{table}[t]
\setlength{\abovecaptionskip}{1mm} 
\setlength{\belowcaptionskip}{-2mm}
\small
\resizebox{1\columnwidth}{!}{
\begin{tabular}{c|cc|cc|cc}
\hline
\multirow{2}{*}{Method} & \multicolumn{2}{c|}{LiTS17} & \multicolumn{2}{c|}{KiTS19} & \multicolumn{2}{c}{AMOS}\\ \cline{2-7}
& DSC($\%$)  & IoU($\%$)  & DSC($\%$)  & IoU($\%$)  & DSC($\%$) & IoU($\%$) \\ \hline\hline
nnUNet~\cite{isensee2021nnu} & 59.77 & 54.77 & 56.65 & 52.49 & 76.14 & 70.01 \\ \hline
MAISI~\cite{guo2024maisi} & 58.47 & 53.43 & 56.73 & 51.65 & 76.56 & 70.28 \\ \hline
ControlNet~\cite{zhang2023adding} & 60.64 &55.36 & 56.88 & 52.34 & \textbf{77.16} & \textbf{70.87} \\ \hline
DiffBoost~\cite{zhang2024diffboost} & 59.92& 55.08 & 56.96 & 52.20 & 76.40 & 70.17 \\ \hline
MedSegFactory  & \textbf{62.37}& \textbf{56.90} & \textbf{58.03} & \textbf{53.39} & 76.76 & 70.48 \\ \hline
\end{tabular}
}
\caption{Segmentation performance evaluation of nnUNet on different combined and real CT datasets.}
\label{tab:seg_ct}
\end{table}

\paragraph{Results.}
Tabs.~\ref{tab:seg_other}--\ref{tab:seg_ct} present the quantitative comparison results between MedSegFactory and other methods for downstream segmentation tasks. Compared to the nnUNet baseline trained exclusively on real datasets, the augmented paired medical images and segmentation labels generated by MedSegFactory significantly improve performance across various medical segmentation tasks. Moreover, compared to other generative models, MedSegFactory most effectively enhances the overall segmentation performance across heterogeneous datasets. These gains highlight its utility in medical segmentation, especially where annotation costs are prohibitive. 

\subsection{Ablation Study}

\begin{table}
\setlength{\abovecaptionskip}{1mm} 
\setlength{\belowcaptionskip}{-5mm}
    \centering  
    \small
    \resizebox{0.9\columnwidth}{!}{
        \footnotesize
        \begin{tabular}{c|cccc}
            \toprule
            \scriptsize{Setting} &  \scriptsize{aFID} & \scriptsize{aIS} & \scriptsize{aDSC} ($\%$) & \scriptsize{aIoU} ($\%$) \\
            \midrule
            \scriptsize{\textit{w/o} JCA} & 143.28 & 2.75 & 61.58 & 57.35 \\ \arrayrulecolor{lightgray}\hline
            \scriptsize{\textit{w/o} mask-to-image branch} & 151.71 & 2.66 & 72.19 & 63.50 \\ \hline
            \scriptsize{\textit{w/o} image-to-mask branch} & 142.64 & 2.70 & 69.83 & 60.72 \\ \hline
            \scriptsize{baseline} & \textbf{138.03} & \textbf{2.83} & \textbf{76.57} & \textbf{71.27}\\  
            \arrayrulecolor{black}\bottomrule
        \end{tabular}
    }
    \caption{\textbf{Ablation study.} JCA and its two internal branches enhance the quality of generated medical images and improve the overall performance across 6 datasets.}
    \label{tab:abla}
\end{table}

\begin{figure}[!t]
    \setlength{\abovecaptionskip}{1mm} 
    \setlength{\belowcaptionskip}{-2mm}
    \centering
    \includegraphics[width=1\linewidth]{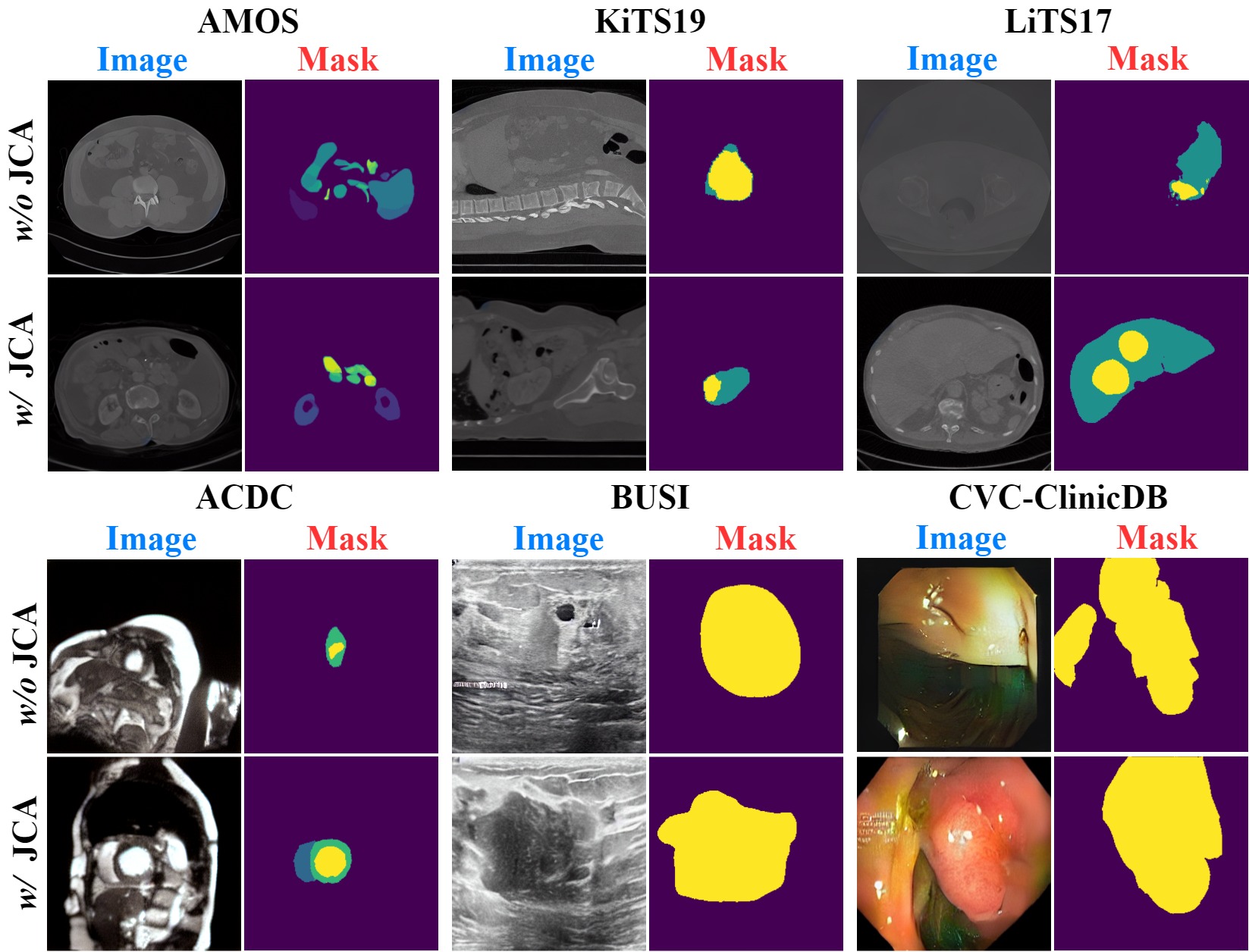}
     \caption{Visualization of the significant misalignment between the generated images and segmentation after removing JCA, along with a noticeable quality decline of the generated images.}
     \label{fig:ablat_jca}
   \end{figure}

We investigate the contribution of JCA and two branches within JCA: the mask-to-image branch and the image-to-mask branch of the MedSegFactory generation framework. We use MedSegFactory as the baseline to evaluate the impact of different ablation settings by reporting 
the average Fréchet Inception Distance (aFID), average Inception Score (aIS), average Dice coefficients (aDSC), and average Intersection over Union (aIoU) scores across all datasets. As shown in Tab.~\ref{tab:abla} and Fig.~\ref{fig:ablat_jca}, JCA successfully enhances the quality of the generated image. Meanwhile, the mask-to-image and image-to-mask branches improve medical image generation and downstream task performance. When both branches are employed, boosting cross-guidance between the generated image features and segmentation features, MedSegFactory demonstrates state-of-the-art performance in both generation and segmentation tasks.

\subsection{Limitation}

\begin{figure}[!t]
    \setlength{\abovecaptionskip}{1mm} 
    \setlength{\belowcaptionskip}{-6mm}
    \centering
    \includegraphics[width=1\linewidth]{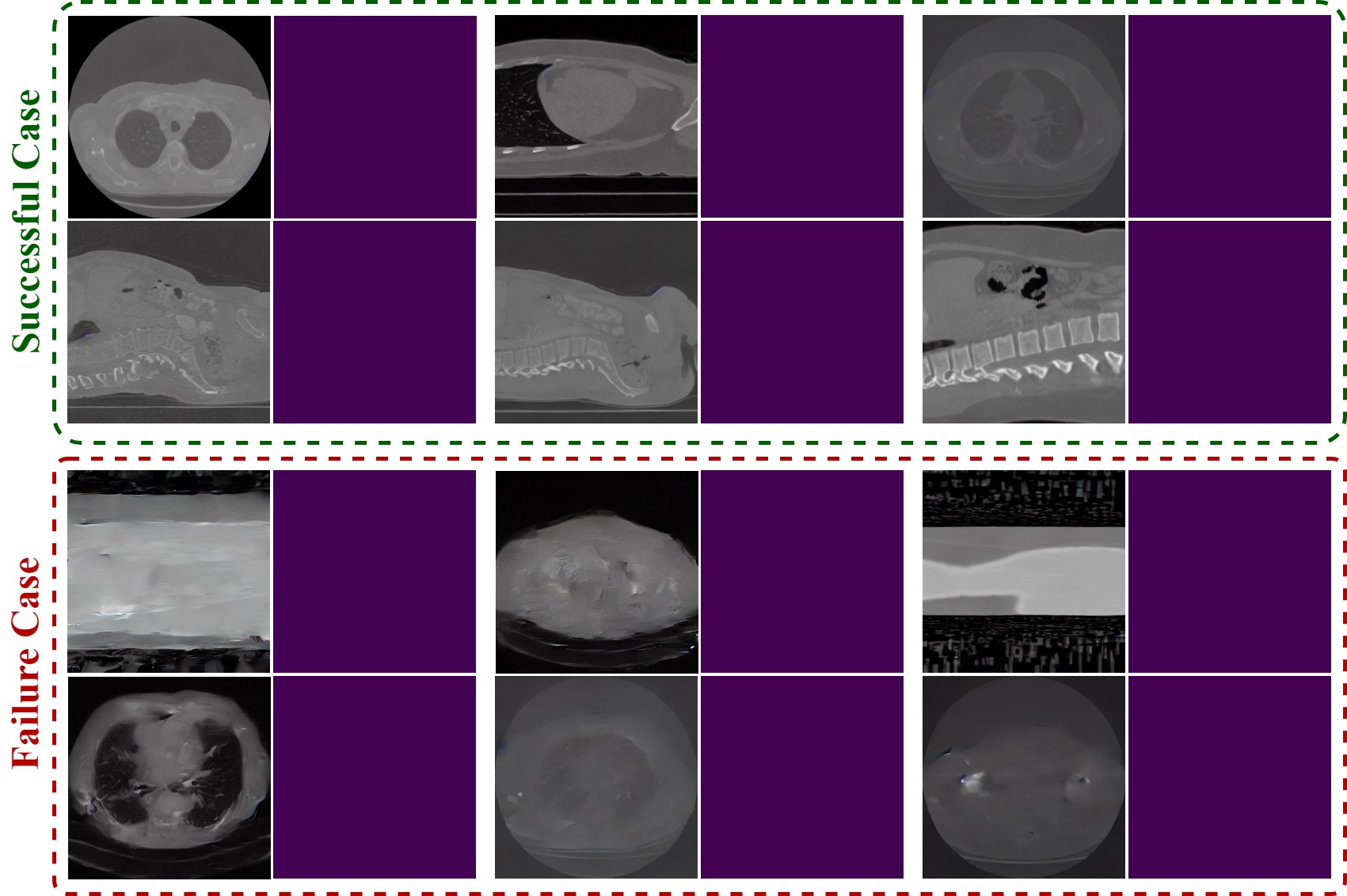}
     \caption{The observed failure cases with blurred boundaries without segmentation targets.}
     \label{fig:failure}
   \end{figure}

\paragraph{Blurred Boundaries in Target-Free Generation.} While MedSegFactory demonstrates promising capabilities in generating paired medical images and segmentation masks, certain failure cases still occur. As shown in Fig.~\ref{fig:failure}, when generating medical images without segmentation targets, the absence of segmentation-guided supervision—an essential element of the JCA—can lead to blurred or imprecise anatomical boundaries. Addressing this limitation presents an opportunity for future improvement. In particular, refining prompt design and introducing more effective attention-masking strategies may help mitigate these issues and enhance generation quality in such cases.

\paragraph{The 2D architecture.} MedSegFactory’s current framework is designed for 2D medical images, which may not fully capture the 3D spatial relationships critical in CT/MRI imaging. Extending the model to 3D volumetric data remains a key direction for future work.
\section{Conclusion}\label{sec:conclusion}
In conclusion, this paper presents MedSegFactory, a novel framework for synthesizing high-quality, diverse, and consistent paired medical images and segmentation masks using concise text prompts. By introducing the Joint Cross-Attention mechanism, MedSegFactory ensures precise alignment and semantic consistency between images and their corresponding segmentation labels. Extensive experiments demonstrate that MedSegFactory outperforms previous state-of-the-art methods in generating medical data and improving segmentation performance across heterogeneous datasets. This work addresses data scarcity and the challenges of regulatory constraints, and provides a valuable tool for advancing computer-aided detection. 

{
    \small
    \bibliographystyle{ieeenat_fullname}
    \bibliography{main}

\begin{thebibliography}{72}
\providecommand{\natexlab}[1]{#1}
\providecommand{\url}[1]{\texttt{#1}}
\expandafter\ifx\csname urlstyle\endcsname\relax
  \providecommand{\doi}[1]{doi: #1}\else
  \providecommand{\doi}{doi: \begingroup \urlstyle{rm}\Url}\fi

\bibitem[Ai et~al.(2024)Ai, Huang, Zhou, Wang, and He]{ai2024multimodal}
Yuang Ai, Huaibo Huang, Xiaoqiang Zhou, Jiexiang Wang, and Ran He.
\newblock Multimodal prompt perceiver: Empower adaptiveness generalizability and fidelity for all-in-one image restoration.
\newblock In \emph{Proceedings of the IEEE/CVF Conference on Computer Vision and Pattern Recognition}, pages 25432--25444, 2024.

\bibitem[Al-Dhabyani et~al.(2020)Al-Dhabyani, Gomaa, Khaled, and Fahmy]{al2020dataset}
Walid Al-Dhabyani, Mohammed Gomaa, Hussien Khaled, and Aly Fahmy.
\newblock Dataset of breast ultrasound images.
\newblock \emph{Data in brief}, 28:\penalty0 104863, 2020.

\bibitem[Azad et~al.(2022)Azad, Heidari, Shariatnia, Aghdam, Karimijafarbigloo, Adeli, and Merhof]{azad2022transdeeplab}
Reza Azad, Moein Heidari, Moein Shariatnia, Ehsan~Khodapanah Aghdam, Sanaz Karimijafarbigloo, Ehsan Adeli, and Dorit Merhof.
\newblock Transdeeplab: Convolution-free transformer-based deeplab v3+ for medical image segmentation.
\newblock In \emph{International Workshop on PRedictive Intelligence In MEdicine}, pages 91--102. Springer, 2022.

\bibitem[Azad et~al.(2023)Azad, Jia, Aghdam, Cohen-Adad, and Merhof]{azad2023enhancing}
Reza Azad, Yiwei Jia, Ehsan~Khodapanah Aghdam, Julien Cohen-Adad, and Dorit Merhof.
\newblock Enhancing medical image segmentation with transception: A multi-scale feature fusion approach.
\newblock \emph{arXiv preprint arXiv:2301.10847}, 2023.

\bibitem[Bernal et~al.(2015)Bernal, S{\'a}nchez, Fern{\'a}ndez-Esparrach, Gil, Rodr{\'\i}guez, and Vilari{\~n}o]{bernal2015wm}
Jorge Bernal, F~Javier S{\'a}nchez, Gloria Fern{\'a}ndez-Esparrach, Debora Gil, Cristina Rodr{\'\i}guez, and Fernando Vilari{\~n}o.
\newblock Wm-dova maps for accurate polyp highlighting in colonoscopy: Validation vs. saliency maps from physicians.
\newblock \emph{Computerized medical imaging and graphics}, 43:\penalty0 99--111, 2015.

\bibitem[Bernard et~al.(2018)Bernard, Lalande, Zotti, Cervenansky, Yang, Heng, Cetin, Lekadir, Camara, Ballester, et~al.]{bernard2018deep}
Olivier Bernard, Alain Lalande, Clement Zotti, Frederick Cervenansky, Xin Yang, Pheng-Ann Heng, Irem Cetin, Karim Lekadir, Oscar Camara, Miguel Angel~Gonzalez Ballester, et~al.
\newblock Deep learning techniques for automatic mri cardiac multi-structures segmentation and diagnosis: is the problem solved?
\newblock \emph{IEEE transactions on medical imaging}, 37\penalty0 (11):\penalty0 2514--2525, 2018.

\bibitem[Bhat et~al.(2025)Bhat, Bose, Nwoye, and Padoy]{bhat2025simgen}
Aditya Bhat, Rupak Bose, Chinedu~Innocent Nwoye, and Nicolas Padoy.
\newblock Simgen: A diffusion-based framework for simultaneous surgical image and segmentation mask generation.
\newblock \emph{arXiv preprint arXiv:2501.09008}, 2025.

\bibitem[Bilic et~al.(2023)Bilic, Christ, Li, Vorontsov, Ben-Cohen, Kaissis, Szeskin, Jacobs, Mamani, Chartrand, et~al.]{bilic2023liver}
Patrick Bilic, Patrick Christ, Hongwei~Bran Li, Eugene Vorontsov, Avi Ben-Cohen, Georgios Kaissis, Adi Szeskin, Colin Jacobs, Gabriel Efrain~Humpire Mamani, Gabriel Chartrand, et~al.
\newblock The liver tumor segmentation benchmark (lits).
\newblock \emph{Medical Image Analysis}, 84:\penalty0 102680, 2023.

\bibitem[Brock(2018)]{brock2018large}
Andrew Brock.
\newblock Large scale gan training for high fidelity natural image synthesis.
\newblock \emph{arXiv preprint arXiv:1809.11096}, 2018.

\bibitem[Cao et~al.(2022)Cao, Wang, Chen, Jiang, Zhang, Tian, and Wang]{cao2022swin}
Hu Cao, Yueyue Wang, Joy Chen, Dongsheng Jiang, Xiaopeng Zhang, Qi Tian, and Manning Wang.
\newblock Swin-unet: Unet-like pure transformer for medical image segmentation.
\newblock In \emph{European conference on computer vision}, pages 205--218. Springer, 2022.

\bibitem[Chen et~al.(2021)Chen, Lu, Yu, Luo, Adeli, Wang, Lu, Yuille, and Zhou]{chen2021transunet}
Jieneng Chen, Yongyi Lu, Qihang Yu, Xiangde Luo, Ehsan Adeli, Yan Wang, Le Lu, Alan~L Yuille, and Yuyin Zhou.
\newblock Transunet: Transformers make strong encoders for medical image segmentation.
\newblock \emph{arXiv preprint arXiv:2102.04306}, 2021.

\bibitem[Chen et~al.(2024{\natexlab{a}})Chen, Chen, Song, Xiong, Yuille, Wei, and Zhou]{chen2024towards}
Qi Chen, Xiaoxi Chen, Haorui Song, Zhiwei Xiong, Alan Yuille, Chen Wei, and Zongwei Zhou.
\newblock Towards generalizable tumor synthesis.
\newblock In \emph{Proceedings of the IEEE/CVF conference on computer vision and pattern recognition}, pages 11147--11158, 2024{\natexlab{a}}.

\bibitem[Chen et~al.(2024{\natexlab{b}})Chen, Lai, Chen, Hu, Yuille, and Zhou]{chen2024analyzing}
Qi Chen, Yuxiang Lai, Xiaoxi Chen, Qixin Hu, Alan Yuille, and Zongwei Zhou.
\newblock Analyzing tumors by synthesis.
\newblock \emph{arXiv preprint arXiv:2409.06035}, 2024{\natexlab{b}}.

\bibitem[Cho et~al.(2024)Cho, Zakka, Shad, Wightman, Chaudhari, and Hiesinger]{cho2024medisyn}
Joseph Cho, Cyril Zakka, Rohan Shad, Ross Wightman, Akshay Chaudhari, and William Hiesinger.
\newblock Medisyn: Text-guided diffusion models for broad medical 2d and 3d image synthesis.
\newblock \emph{arXiv e-prints}, pages arXiv--2405, 2024.

\bibitem[Dalmaz et~al.(2022)Dalmaz, Yurt, and {\c{C}}ukur]{dalmaz2022resvit}
Onat Dalmaz, Mahmut Yurt, and Tolga {\c{C}}ukur.
\newblock Resvit: residual vision transformers for multimodal medical image synthesis.
\newblock \emph{IEEE Transactions on Medical Imaging}, 41\penalty0 (10):\penalty0 2598--2614, 2022.

\bibitem[Dorjsembe et~al.(2024{\natexlab{a}})Dorjsembe, Pao, Odonchimed, and Xiao]{dorjsembe2024conditional}
Zolnamar Dorjsembe, Hsing-Kuo Pao, Sodtavilan Odonchimed, and Furen Xiao.
\newblock Conditional diffusion models for semantic 3d brain mri synthesis.
\newblock \emph{IEEE Journal of Biomedical and Health Informatics}, 2024{\natexlab{a}}.

\bibitem[Dorjsembe et~al.(2024{\natexlab{b}})Dorjsembe, Pao, and Xiao]{dorjsembe2024polyp}
Zolnamar Dorjsembe, Hsing-Kuo Pao, and Furen Xiao.
\newblock Polyp-ddpm: Diffusion-based semantic polyp synthesis for enhanced segmentation.
\newblock In \emph{2024 46th Annual International Conference of the IEEE Engineering in Medicine and Biology Society (EMBC)}, pages 1--7. IEEE, 2024{\natexlab{b}}.

\bibitem[Du et~al.(2023)Du, Jiang, Tan, Wu, Dou, Li, Li, and Wan]{du2023arsdm}
Yuhao Du, Yuncheng Jiang, Shuangyi Tan, Xusheng Wu, Qi Dou, Zhen Li, Guanbin Li, and Xiang Wan.
\newblock Arsdm: Colonoscopy images synthesis with adaptive refinement semantic diffusion models.
\newblock In \emph{International Conference on Medical Image Computing and Computer-Assisted Intervention}, pages 339--349. Springer, 2023.

\bibitem[Esser et~al.(2023)Esser, Chiu, Atighehchian, Granskog, and Germanidis]{esser2023structure}
Patrick Esser, Johnathan Chiu, Parmida Atighehchian, Jonathan Granskog, and Anastasis Germanidis.
\newblock Structure and content-guided video synthesis with diffusion models.
\newblock In \emph{Proceedings of the IEEE/CVF International Conference on Computer Vision}, pages 7346--7356, 2023.

\bibitem[Friedrich et~al.(2024)Friedrich, Wolleb, Bieder, Durrer, and Cattin]{friedrich2024wdm}
Paul Friedrich, Julia Wolleb, Florentin Bieder, Alicia Durrer, and Philippe~C Cattin.
\newblock Wdm: 3d wavelet diffusion models for high-resolution medical image synthesis.
\newblock In \emph{MICCAI Workshop on Deep Generative Models}, pages 11--21. Springer, 2024.

\bibitem[Goodfellow et~al.(2020)Goodfellow, Pouget-Abadie, Mirza, Xu, Warde-Farley, Ozair, Courville, and Bengio]{goodfellow2020generative}
Ian Goodfellow, Jean Pouget-Abadie, Mehdi Mirza, Bing Xu, David Warde-Farley, Sherjil Ozair, Aaron Courville, and Yoshua Bengio.
\newblock Generative adversarial networks.
\newblock \emph{Communications of the ACM}, 63\penalty0 (11):\penalty0 139--144, 2020.

\bibitem[Guo et~al.(2024)Guo, Zhao, Yang, Xu, Nath, Tang, Simon, Belue, Harmon, Turkbey, et~al.]{guo2024maisi}
Pengfei Guo, Can Zhao, Dong Yang, Ziyue Xu, Vishwesh Nath, Yucheng Tang, Benjamin Simon, Mason Belue, Stephanie Harmon, Baris Turkbey, et~al.
\newblock Maisi: Medical ai for synthetic imaging.
\newblock \emph{arXiv preprint arXiv:2409.11169}, 2024.

\bibitem[Heller et~al.(2021)Heller, Isensee, Maier-Hein, Hou, Xie, Li, Nan, Mu, Lin, Han, Yao, Gao, Zhang, Wang, Hou, Yang, Xiong, Tian, Zhong, Ma, Rickman, Dean, Stai, Tejpaul, Oestreich, Blake, Kaluzniak, Raza, Rosenberg, Moore, Walczak, Rengel, Edgerton, Vasdev, Peterson, McSweeney, Peterson, Kalapara, Sathianathen, Papanikolopoulos, and Weight]{heller2023kits19}
Nicholas Heller, Fabian Isensee, Klaus~H. Maier-Hein, Xiaoshuai Hou, Chunmei Xie, Fengyi Li, Yang Nan, Guangrui Mu, Zhiyong Lin, Miofei Han, Guang Yao, Yaozong Gao, Yao Zhang, Yixin Wang, Feng Hou, Jiawei Yang, Guangwei Xiong, Jiang Tian, Cheng Zhong, Jun Ma, Jack Rickman, Joshua Dean, Bethany Stai, Resha Tejpaul, Makinna Oestreich, Paul Blake, Heather Kaluzniak, Shaneabbas Raza, Joel Rosenberg, Keenan Moore, Edward Walczak, Zachary Rengel, Zach Edgerton, Ranveer Vasdev, Matthew Peterson, Sean McSweeney, Sarah Peterson, Arveen Kalapara, Niranjan Sathianathen, Nikolaos Papanikolopoulos, and Christopher Weight.
\newblock The state of the art in kidney and kidney tumor segmentation in contrast-enhanced ct imaging: Results of the kits19 challenge.
\newblock \emph{Medical Image Analysis}, 67:\penalty0 101821, 2021.

\bibitem[Ho et~al.(2020)Ho, Jain, and Abbeel]{ho2020denoising}
Jonathan Ho, Ajay Jain, and Pieter Abbeel.
\newblock Denoising diffusion probabilistic models.
\newblock \emph{Advances in neural information processing systems}, 33:\penalty0 6840--6851, 2020.

\bibitem[Huang et~al.(2020)Huang, Lin, Tong, Hu, Zhang, Iwamoto, Han, Chen, and Wu]{huang2020unet}
Huimin Huang, Lanfen Lin, Ruofeng Tong, Hongjie Hu, Qiaowei Zhang, Yutaro Iwamoto, Xianhua Han, Yen-Wei Chen, and Jian Wu.
\newblock Unet 3+: A full-scale connected unet for medical image segmentation.
\newblock In \emph{ICASSP 2020-2020 IEEE international conference on acoustics, speech and signal processing (ICASSP)}, pages 1055--1059. IEEE, 2020.

\bibitem[Huang et~al.(2021)Huang, Deng, Li, and Yuan]{huang2021missformer}
Xiaohong Huang, Zhifang Deng, Dandan Li, and Xueguang Yuan.
\newblock Missformer: An effective medical image segmentation transformer.
\newblock \emph{arXiv preprint arXiv:2109.07162}, 2021.

\bibitem[Isensee et~al.(2021)Isensee, Jaeger, Kohl, Petersen, and Maier-Hein]{isensee2021nnu}
Fabian Isensee, Paul~F Jaeger, Simon~AA Kohl, Jens Petersen, and Klaus~H Maier-Hein.
\newblock nnu-net: a self-configuring method for deep learning-based biomedical image segmentation.
\newblock \emph{Nature methods}, 18\penalty0 (2):\penalty0 203--211, 2021.

\bibitem[Jha et~al.(2020)Jha, Smedsrud, Riegler, Halvorsen, de~Lange, Johansen, and Johansen]{jha2020kvasir}
Debesh Jha, Pia~H Smedsrud, Michael~A Riegler, P{\aa}l Halvorsen, Thomas de Lange, Dag Johansen, and H{\aa}vard~D Johansen.
\newblock Kvasir-seg: A segmented polyp dataset.
\newblock In \emph{MultiMedia Modeling: 26th International Conference, MMM 2020, Daejeon, South Korea, January 5--8, 2020, Proceedings, Part II 26}, pages 451--462. Springer, 2020.

\bibitem[Ji et~al.(2022)Ji, Bai, Ge, Yang, Zhu, Zhang, Li, Zhanng, Ma, Wan, et~al.]{ji2022amos}
Yuanfeng Ji, Haotian Bai, Chongjian Ge, Jie Yang, Ye Zhu, Ruimao Zhang, Zhen Li, Lingyan Zhanng, Wanling Ma, Xiang Wan, et~al.
\newblock Amos: A large-scale abdominal multi-organ benchmark for versatile medical image segmentation.
\newblock \emph{Advances in neural information processing systems}, 35:\penalty0 36722--36732, 2022.

\bibitem[Jiang et~al.(2025)Jiang, Lemar{\'e}chal, Bafaro, Abi-Rjeile, Joubert, Despr{\'e}s, and Manem]{jiang2025lung}
Yifan Jiang, Yannick Lemar{\'e}chal, Jos{\'e}e Bafaro, Jessica Abi-Rjeile, Philippe Joubert, Philippe Despr{\'e}s, and Venkata Manem.
\newblock Lung-ddpm: Semantic layout-guided diffusion models for thoracic ct image synthesis.
\newblock \emph{arXiv preprint arXiv:2502.15204}, 2025.

\bibitem[Kang et~al.(2023)Kang, Li, Zhu, Lu, Fishman, Yuille, and Zhou]{kang2023label}
Mintong Kang, Bowen Li, Zengle Zhu, Yongyi Lu, Elliot~K Fishman, Alan Yuille, and Zongwei Zhou.
\newblock Label-assemble: Leveraging multiple datasets with partial labels.
\newblock In \emph{2023 IEEE 20th International Symposium on Biomedical Imaging (ISBI)}, pages 1--5. IEEE, 2023.

\bibitem[Karras et~al.(2020)Karras, Laine, Aittala, Hellsten, Lehtinen, and Aila]{Karras2019stylegan2}
Tero Karras, Samuli Laine, Miika Aittala, Janne Hellsten, Jaakko Lehtinen, and Timo Aila.
\newblock Analyzing and improving the image quality of {StyleGAN}.
\newblock In \emph{Proc. CVPR}, 2020.

\bibitem[Khachatryan et~al.(2023)Khachatryan, Movsisyan, Tadevosyan, Henschel, Wang, Navasardyan, and Shi]{khachatryan2023text2video}
Levon Khachatryan, Andranik Movsisyan, Vahram Tadevosyan, Roberto Henschel, Zhangyang Wang, Shant Navasardyan, and Humphrey Shi.
\newblock Text2video-zero: Text-to-image diffusion models are zero-shot video generators.
\newblock In \emph{Proceedings of the IEEE/CVF International Conference on Computer Vision}, pages 15954--15964, 2023.

\bibitem[Kirillov et~al.(2023)Kirillov, Mintun, Ravi, Mao, Rolland, Gustafson, Xiao, Whitehead, Berg, Lo, et~al.]{kirillov2023segment}
Alexander Kirillov, Eric Mintun, Nikhila Ravi, Hanzi Mao, Chloe Rolland, Laura Gustafson, Tete Xiao, Spencer Whitehead, Alexander~C Berg, Wan-Yen Lo, et~al.
\newblock Segment anything.
\newblock In \emph{Proceedings of the IEEE/CVF international conference on computer vision}, pages 4015--4026, 2023.

\bibitem[Lee et~al.(2024)Lee, Kwon, and Kim]{lee2024grid}
Taegyeong Lee, Soyeong Kwon, and Taehwan Kim.
\newblock Grid diffusion models for text-to-video generation.
\newblock In \emph{Proceedings of the IEEE/CVF Conference on Computer Vision and Pattern Recognition}, pages 8734--8743, 2024.

\bibitem[Li et~al.(2024{\natexlab{a}})Li, Qu, Chen, Bassi, Shi, Lai, Yu, Xue, Chen, Lin, et~al.]{li2024abdomenatlas}
Wenxuan Li, Chongyu Qu, Xiaoxi Chen, Pedro~RAS Bassi, Yijia Shi, Yuxiang Lai, Qian Yu, Huimin Xue, Yixiong Chen, Xiaorui Lin, et~al.
\newblock Abdomenatlas: A large-scale, detailed-annotated, \& multi-center dataset for efficient transfer learning and open algorithmic benchmarking.
\newblock \emph{Medical Image Analysis}, page 103285, 2024{\natexlab{a}}.

\bibitem[Li et~al.(2024{\natexlab{b}})Li, Shuai, Liu, Chen, Wu, Guo, Yang, Zhao, Bassi, Xu, et~al.]{li2024text}
Xinran Li, Yi Shuai, Chen Liu, Qi Chen, Qilong Wu, Pengfei Guo, Dong Yang, Can Zhao, Pedro~RAS Bassi, Daguang Xu, et~al.
\newblock Text-driven tumor synthesis.
\newblock \emph{arXiv preprint arXiv:2412.18589}, 2024{\natexlab{b}}.

\bibitem[Liu et~al.(2023)Liu, Zhang, Chen, Xiao, Lu, A~Landman, Yuan, Yuille, Tang, and Zhou]{liu2023clip}
Jie Liu, Yixiao Zhang, Jie-Neng Chen, Junfei Xiao, Yongyi Lu, Bennett A~Landman, Yixuan Yuan, Alan Yuille, Yucheng Tang, and Zongwei Zhou.
\newblock Clip-driven universal model for organ segmentation and tumor detection.
\newblock In \emph{Proceedings of the IEEE/CVF international conference on computer vision}, pages 21152--21164, 2023.

\bibitem[Long et~al.(2015)Long, Shelhamer, and Darrell]{long2015fully}
Jonathan Long, Evan Shelhamer, and Trevor Darrell.
\newblock Fully convolutional networks for semantic segmentation.
\newblock In \emph{Proceedings of the IEEE conference on computer vision and pattern recognition}, pages 3431--3440, 2015.

\bibitem[Luo et~al.(2023)Luo, Tan, Huang, Li, and Zhao]{luo2023latent}
Simian Luo, Yiqin Tan, Longbo Huang, Jian Li, and Hang Zhao.
\newblock Latent consistency models: Synthesizing high-resolution images with few-step inference, 2023.

\bibitem[Ma et~al.(2024)Ma, He, Li, Han, You, and Wang]{ma2024segment}
Jun Ma, Yuting He, Feifei Li, Lin Han, Chenyu You, and Bo Wang.
\newblock Segment anything in medical images.
\newblock \emph{Nature Communications}, 15\penalty0 (1):\penalty0 654, 2024.

\bibitem[Mach{\'a}{\v{c}}ek et~al.(2023)Mach{\'a}{\v{c}}ek, Mozaffari, Sepasdar, Parasa, Halvorsen, Riegler, and Thambawita]{machavcek2023mask}
Roman Mach{\'a}{\v{c}}ek, Leila Mozaffari, Zahra Sepasdar, Sravanthi Parasa, P{\aa}l Halvorsen, Michael~A Riegler, and Vajira Thambawita.
\newblock Mask-conditioned latent diffusion for generating gastrointestinal polyp images.
\newblock In \emph{Proceedings of the 4th ACM Workshop on Intelligent Cross-Data Analysis and Retrieval}, pages 1--9, 2023.

\bibitem[Nichol and Dhariwal(2021)]{nichol2021improved}
Alexander~Quinn Nichol and Prafulla Dhariwal.
\newblock Improved denoising diffusion probabilistic models.
\newblock In \emph{International conference on machine learning}, pages 8162--8171. PMLR, 2021.

\bibitem[Nie et~al.(2018)Nie, Trullo, Lian, Wang, Petitjean, Ruan, Wang, and Shen]{nie2018medical}
Dong Nie, Roger Trullo, Jun Lian, Li Wang, Caroline Petitjean, Su Ruan, Qian Wang, and Dinggang Shen.
\newblock Medical image synthesis with deep convolutional adversarial networks.
\newblock \emph{IEEE Transactions on Biomedical Engineering}, 65\penalty0 (12):\penalty0 2720--2730, 2018.

\bibitem[Perera et~al.(2024)Perera, Navard, and Yilmaz]{perera2024segformer3d}
Shehan Perera, Pouyan Navard, and Alper Yilmaz.
\newblock Segformer3d: an efficient transformer for 3d medical image segmentation.
\newblock In \emph{Proceedings of the IEEE/CVF Conference on Computer Vision and Pattern Recognition}, pages 4981--4988, 2024.

\bibitem[Podell et~al.(2023)Podell, English, Lacey, Blattmann, Dockhorn, M{\"u}ller, Penna, and Rombach]{podell2023sdxl}
Dustin Podell, Zion English, Kyle Lacey, Andreas Blattmann, Tim Dockhorn, Jonas M{\"u}ller, Joe Penna, and Robin Rombach.
\newblock Sdxl: Improving latent diffusion models for high-resolution image synthesis.
\newblock \emph{arXiv preprint arXiv:2307.01952}, 2023.

\bibitem[Polamreddy et~al.(2024)Polamreddy, Roy, Yueh, Mahato, Kuppili, Li, and Zhang]{polamreddy2024leapfrog}
Lakshmikar~R Polamreddy, Kalyan Roy, Sheng-Han Yueh, Deepshikha Mahato, Shilpa Kuppili, Jialu Li, and Youshan Zhang.
\newblock Leapfrog latent consistency model (llcm) for medical images generation.
\newblock \emph{arXiv preprint arXiv:2411.15084}, 2024.

\bibitem[Qadir et~al.(2022)Qadir, Balasingham, and Shin]{qadir2022simple}
Hemin~Ali Qadir, Ilangko Balasingham, and Younghak Shin.
\newblock Simple u-net based synthetic polyp image generation: Polyp to negative and negative to polyp.
\newblock \emph{Biomedical Signal Processing and Control}, 74:\penalty0 103491, 2022.

\bibitem[Ren et~al.(2024)Ren, Huang, Li, Xiao, Mei, Wang, Yuille, and Zhou]{ren2024medical}
Sucheng Ren, Xiaoke Huang, Xianhang Li, Junfei Xiao, Jieru Mei, Zeyu Wang, Alan Yuille, and Yuyin Zhou.
\newblock Medical vision generalist: Unifying medical imaging tasks in context.
\newblock \emph{arXiv preprint arXiv:2406.05565}, 2024.

\bibitem[Rombach et~al.(2022)Rombach, Blattmann, Lorenz, Esser, and Ommer]{rombach2022high}
Robin Rombach, Andreas Blattmann, Dominik Lorenz, Patrick Esser, and Bj{\"o}rn Ommer.
\newblock High-resolution image synthesis with latent diffusion models.
\newblock In \emph{Proceedings of the IEEE/CVF conference on computer vision and pattern recognition}, pages 10684--10695, 2022.

\bibitem[Ronneberger et~al.(2015)Ronneberger, Fischer, and Brox]{ronneberger2015u}
Olaf Ronneberger, Philipp Fischer, and Thomas Brox.
\newblock U-net: Convolutional networks for biomedical image segmentation.
\newblock In \emph{Medical image computing and computer-assisted intervention--MICCAI 2015: 18th international conference, Munich, Germany, October 5-9, 2015, proceedings, part III 18}, pages 234--241. Springer, 2015.

\bibitem[Ruan et~al.(2024)Ruan, Li, and Xiang]{ruan2024vm}
Jiacheng Ruan, Jincheng Li, and Suncheng Xiang.
\newblock Vm-unet: Vision mamba unet for medical image segmentation.
\newblock \emph{arXiv preprint arXiv:2402.02491}, 2024.

\bibitem[Sams and Shomee(2022)]{sams2022gan}
Ataher Sams and Homaira~Huda Shomee.
\newblock Gan-based realistic gastrointestinal polyp image synthesis.
\newblock In \emph{2022 IEEE 19th International Symposium on Biomedical Imaging (ISBI)}, pages 1--4. IEEE, 2022.

\bibitem[Schlemper et~al.(2019)Schlemper, Oktay, Schaap, Heinrich, Kainz, Glocker, and Rueckert]{schlemper2019attention}
Jo Schlemper, Ozan Oktay, Michiel Schaap, Mattias Heinrich, Bernhard Kainz, Ben Glocker, and Daniel Rueckert.
\newblock Attention gated networks: Learning to leverage salient regions in medical images.
\newblock \emph{Medical image analysis}, 53:\penalty0 197--207, 2019.

\bibitem[Skandarani et~al.(2023)Skandarani, Jodoin, and Lalande]{skandarani2023gans}
Youssef Skandarani, Pierre-Marc Jodoin, and Alain Lalande.
\newblock Gans for medical image synthesis: An empirical study.
\newblock \emph{Journal of Imaging}, 9\penalty0 (3):\penalty0 69, 2023.

\bibitem[Song et~al.(2020{\natexlab{a}})Song, Meng, and Ermon]{song2020denoising}
Jiaming Song, Chenlin Meng, and Stefano Ermon.
\newblock Denoising diffusion implicit models.
\newblock \emph{arXiv preprint arXiv:2010.02502}, 2020{\natexlab{a}}.

\bibitem[Song et~al.(2020{\natexlab{b}})Song, Sohl-Dickstein, Kingma, Kumar, Ermon, and Poole]{song2020score}
Yang Song, Jascha Sohl-Dickstein, Diederik~P Kingma, Abhishek Kumar, Stefano Ermon, and Ben Poole.
\newblock Score-based generative modeling through stochastic differential equations.
\newblock \emph{arXiv preprint arXiv:2011.13456}, 2020{\natexlab{b}}.

\bibitem[Sun et~al.(2022)Sun, Chen, Xu, Gong, Yu, and Batmanghelich]{hagan2022}
Li Sun, Junxiang Chen, Yanwu Xu, Mingming Gong, Ke Yu, and Kayhan Batmanghelich.
\newblock Hierarchical amortized gan for 3d high resolution medical image synthesis.
\newblock \emph{IEEE Journal of Biomedical and Health Informatics}, 26\penalty0 (8):\penalty0 3966--3975, 2022.

\bibitem[Thambawita et~al.(2022)Thambawita, Salehi, Sheshkal, Hicks, Hammer, Parasa, Lange, Halvorsen, and Riegler]{thambawita2022singan}
Vajira Thambawita, Pegah Salehi, Sajad~Amouei Sheshkal, Steven~A Hicks, Hugo~L Hammer, Sravanthi Parasa, Thomas~de Lange, P{\aa}l Halvorsen, and Michael~A Riegler.
\newblock Singan-seg: Synthetic training data generation for medical image segmentation.
\newblock \emph{PloS one}, 17\penalty0 (5):\penalty0 e0267976, 2022.

\bibitem[Wang and Hu(2024)]{wang2024image}
Jiarong Wang and Hao Hu.
\newblock Image dehazing based on iterative-refining diffusion model.
\newblock In \emph{Proceedings of the 2024 7th International Conference on Image and Graphics Processing}, pages 355--362, 2024.

\bibitem[Wang et~al.(2024)Wang, Liu, Su, Yang, and Gao]{wang2024fsam}
Xiaoming Wang, Lei Liu, Xiangdong Su, Yuhan Yang, and Guanglai Gao.
\newblock Fsam: Fine-tuning sam encoder and decoder for medical image segmentation.
\newblock In \emph{2024 IEEE International Conference on Bioinformatics and Biomedicine (BIBM)}, pages 5569--5573. IEEE, 2024.

\bibitem[Wu et~al.(2024)Wu, Zhao, Zhang, Xie, and Wang]{wu2024mrgen}
Haoning Wu, Ziheng Zhao, Ya Zhang, Weidi Xie, and Yanfeng Wang.
\newblock Mrgen: Diffusion-based controllable data engine for mri segmentation towards unannotated modalities.
\newblock \emph{arXiv preprint arXiv:2412.04106}, 2024.

\bibitem[Yan et~al.(2022)Yan, Tang, Sun, Ma, Kong, and Xie]{yan2022after}
Xiangyi Yan, Hao Tang, Shanlin Sun, Haoyu Ma, Deying Kong, and Xiaohui Xie.
\newblock After-unet: Axial fusion transformer unet for medical image segmentation.
\newblock In \emph{Proceedings of the IEEE/CVF winter conference on applications of computer vision}, pages 3971--3981, 2022.

\bibitem[Yi et~al.(2023)Yi, Xu, Zhang, Tang, and Ma]{yi2023diff}
Xunpeng Yi, Han Xu, Hao Zhang, Linfeng Tang, and Jiayi Ma.
\newblock Diff-retinex: Rethinking low-light image enhancement with a generative diffusion model.
\newblock In \emph{Proceedings of the IEEE/CVF International Conference on Computer Vision}, pages 12302--12311, 2023.

\bibitem[Zhang et~al.(2023{\natexlab{a}})Zhang, Rao, and Agrawala]{zhang2023adding}
Lvmin Zhang, Anyi Rao, and Maneesh Agrawala.
\newblock Adding conditional control to text-to-image diffusion models, 2023{\natexlab{a}}.

\bibitem[Zhang et~al.(2023{\natexlab{b}})Zhang, Xie, Huang, Zhang, Chen, Tian, and Wang]{zhang2023self}
Xiaoman Zhang, Weidi Xie, Chaoqin Huang, Ya Zhang, Xin Chen, Qi Tian, and Yanfeng Wang.
\newblock Self-supervised tumor segmentation with sim2real adaptation.
\newblock \emph{IEEE Journal of Biomedical and Health Informatics}, 27\penalty0 (9):\penalty0 4373--4384, 2023{\natexlab{b}}.

\bibitem[Zhang et~al.(2021)Zhang, Liu, and Hu]{zhang2021transfuse}
Yundong Zhang, Huiye Liu, and Qiang Hu.
\newblock Transfuse: Fusing transformers and cnns for medical image segmentation.
\newblock \emph{arXiv preprint arXiv:2102.08005}, 2021.

\bibitem[Zhang et~al.(2024)Zhang, Yao, Wang, Jha, Durak, Keles, Medetalibeyoglu, and Bagci]{zhang2024diffboost}
Zheyuan Zhang, Lanhong Yao, Bin Wang, Debesh Jha, Gorkem Durak, Elif Keles, Alpay Medetalibeyoglu, and Ulas Bagci.
\newblock Diffboost: Enhancing medical image segmentation via text-guided diffusion model.
\newblock \emph{IEEE Transactions on Medical Imaging}, 2024.

\bibitem[Zhou et~al.(2019{\natexlab{a}})Zhou, He, Cui, Zhu, Liu, and Shao]{zhou2019high}
Yi Zhou, Xiaodong He, Shanshan Cui, Fan Zhu, Li Liu, and Ling Shao.
\newblock High-resolution diabetic retinopathy image synthesis manipulated by grading and lesions.
\newblock In \emph{International conference on medical image computing and computer-assisted intervention}, pages 505--513. Springer, 2019{\natexlab{a}}.

\bibitem[Zhou et~al.(2019{\natexlab{b}})Zhou, Li, Bai, Wang, Chen, Han, Fishman, and Yuille]{zhou2019prior}
Yuyin Zhou, Zhe Li, Song Bai, Chong Wang, Xinlei Chen, Mei Han, Elliot Fishman, and Alan~L Yuille.
\newblock Prior-aware neural network for partially-supervised multi-organ segmentation.
\newblock In \emph{Proceedings of the IEEE/CVF international conference on computer vision}, pages 10672--10681, 2019{\natexlab{b}}.

\bibitem[Zhou et~al.(2019{\natexlab{c}})Zhou, Wang, Tang, Bai, Shen, Fishman, and Yuille]{zhou2019semi}
Yuyin Zhou, Yan Wang, Peng Tang, Song Bai, Wei Shen, Elliot Fishman, and Alan Yuille.
\newblock Semi-supervised 3d abdominal multi-organ segmentation via deep multi-planar co-training.
\newblock In \emph{2019 IEEE Winter Conference on Applications of Computer Vision (WACV)}, pages 121--140. IEEE, 2019{\natexlab{c}}.

\bibitem[Zhu et~al.(2017)Zhu, Park, Isola, and Efros]{zhu2017unpaired}
Jun-Yan Zhu, Taesung Park, Phillip Isola, and Alexei~A Efros.
\newblock Unpaired image-to-image translation using cycle-consistent adversarial networks.
\newblock In \emph{Proceedings of the IEEE international conference on computer vision}, pages 2223--2232, 2017.

\end{thebibliography}
}

\end{document}